\pdfoutput=1

\documentclass[11pt]{article}

\usepackage[]{EMNLP2022}

\usepackage{times}
\usepackage{latexsym}

\usepackage[T1]{fontenc}

\usepackage[utf8]{inputenc}

\usepackage{amsfonts}
\usepackage{amsmath}
\usepackage{graphicx}
\usepackage{multirow}
\usepackage{makecell}
\usepackage{caption}
\usepackage{subfigure}
\usepackage{enumitem}
\usepackage{float}
\usepackage{makecell}
\usepackage{booktabs}
\usepackage{tabu}
\usepackage{tabularx}
\usepackage{pifont}

\newcommand{\sub}{\textsubscript}

\title{Style Transfer as Data Augmentation: A Case Study \\ on Named Entity Recognition}

\author{
    Shuguang Chen \\ University of Houston \\ \texttt{schen52@uh.edu} \\
    \And
    Leonardo Neves \\ Snap Inc. \\ \texttt{lneves@snap.com} \\
    \And
    Thamar Solorio \\ University of Houston \\ \texttt{tsolorio@uh.edu} \\
}

\begin{document}
\maketitle

\begin{abstract}
In this work, we take the named entity recognition task in the English language as a case study and explore style transfer as a data augmentation method to increase the size and diversity of training data in low-resource scenarios. We propose a new method to effectively transform the text from a high-resource domain to a low-resource domain by changing its style-related attributes to generate synthetic data for training. Moreover, we design a constrained decoding algorithm along with a set of key ingredients for data selection to guarantee the generation of valid and coherent data. Experiments and analysis on five different domain pairs under different data regimes demonstrate that our approach can significantly improve results compared to current state-of-the-art data augmentation methods. Our approach is a practical solution to data scarcity, and we expect it to be applicable to other NLP tasks. \footnote{We released the code at \url{https://github.com/RiTUAL-UH/DA_NER}.}
\end{abstract}

\section{Introduction}
Large-scale pre-trained language models (PLMs) such as BERT \citep{devlin-etal-2019-bert} and T5 \citep{2020t5} have shown impressive performances on a wide variety of NLP tasks. Following the paradigm of self-supervised pre-training and fine-tuning, these models have achieved state-of-the-art performance in many NLP benchmarks such as question answering \citep{DBLP:conf/nips/YangDYCSL19, yamada-etal-2020-luke}, machine translation \citep{provilkov-etal-2020-bpe, lewis-etal-2020-bart}, and text summarization \citep{DBLP:conf/nips/ZaheerGDAAOPRWY20, DBLP:conf/iclr/AghajanyanSGGZG21}. However, due to the discrepancy of training objectives between language modeling and downstream tasks, the performance of such models may be limited by data scarcity in low-resource domains \citep{jiang-etal-2020-smart, gururangan-etal-2020-dont}. 

Data augmentation is effective in addressing data scarcity. Previous work \citep{wei-zou-2019-eda, morris-etal-2020-textattack} has mainly focused on using in-domain data to synthesize new datasets for training. When applied to low-resource domains, however, these approaches may not lead to significant improvement gains as the data in low-resource domains is not comparable to that in high-resource domains in terms of size and diversity. Recently, many studies \citep{xia-etal-2019-generalized, dehouck-gomez-rodriguez-2020-data} have revealed the potential of leveraging data from high-resource domains to improve low-resource tasks. Despite their impressive results, directly using abundant data from high-resource domains can be problematic due to the difference in data distribution (e.g., language shift) and feature misalignment (e.g., class mismatch) between domains \citep{wang-etal-2018-label-aware, zhang-etal-2021-pdaln}.

\begin{figure*}[ht]
\centering
\includegraphics[width=0.95\linewidth]{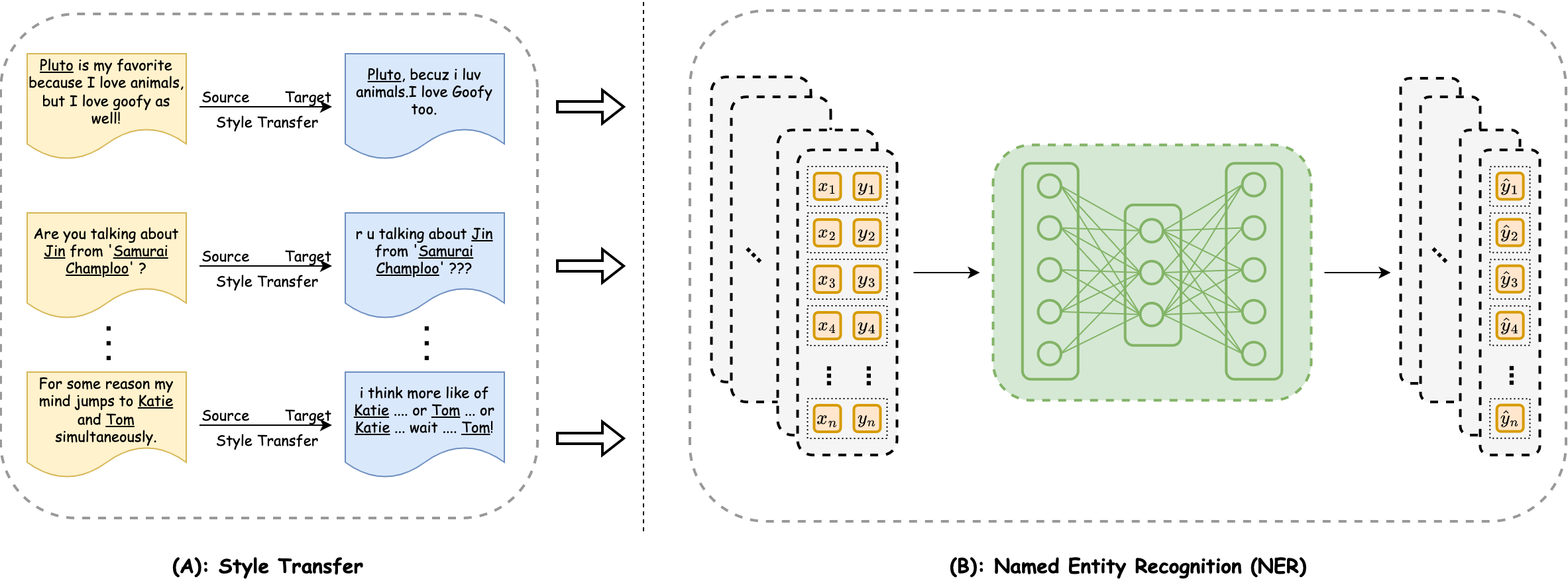}
\caption{The general pipeline of employing style transfer as data augmentation. For each sample from the source domain, we transfer it to the target domain by changing its style-related attributes as synthetic data, and then use it to train NER models. In the figure, the source and target domains are newswire and social media, respectively, and the underline words are entity instances.}
\label{fig: pipeline}
\end{figure*}

In this work, we explore the potential of employing style transfer as a way of data augmentation in cross-domain settings. Style transfer on natural language aims to change the style-related attributes of text while preserving its semantics, which makes it a reasonable alternative for the purpose of data augmentation. Here, we take the named entity recognition (NER) task as a case study to investigate its effectiveness. The general pipeline is shown in Figure \ref{fig: pipeline}. Compared to the text classification task, the NER task is more difficult as it requires to jointly modify the tokens and their corresponding labels. One of the critical challenges here is the lack of parallel style transfer data annotated with NER labels. Our workaround to the lack of data is to figure out how to leverage a non-NER parallel style transfer dataset and a nonparallel NER dataset. Both datasets contain pairs of sentences in source and target styles, respectively. This scenario is much more realistic since resources for style transfer tend to be task-agnostic. At the same time, it is easier to come by nonparallel task-specific datasets with different styles. Moreover, a solution that can successfully exploit data in this manner will be relevant in general data augmentation scenarios beyond NER.

We formulate style transfer as a paraphrase generation problem following previous work \citep{krishna-etal-2020-reformulating, qi-etal-2021-mind} and propose a novel neural architecture that uses PLMs as the generator and discriminator. The generator aims to transform text guided by task prefixes while the discriminator aims to differentiate text between different styles. We jointly train the generator and discriminator on both parallel and nonparallel data to learn transformations in a semi-supervised manner. We apply paraphrase generation directly on the parallel data to bridge the gap between source and target styles. For nonparallel data, we use a cycle-consistent reconstruction to re-paraphrase back the paraphrased sentences to its original style. Additionally, to guarantee the generation of valid and coherent data, we present a constrained decoding algorithm based on prefix trees along with a set of key ingredients for data selection to reduce the noise in synthetic data. We systematically evaluate our proposed methods on 5 different domain pairs under different data regimes. Experiments and analysis show that our proposed method can effectively generate synthetic data by imitating the style of a target low-resource domain and significantly outperform previous state-of-the-art methods.

In summary, our contributions are as follows:
\begin{enumerate}[topsep=0pt,itemsep=-1ex,partopsep=1ex,parsep=1ex]
\item We propose a novel approach for data augmentation that leverages style transfer to improve the low-resource NER task and show that our approach can consistently outperform previous state-of-the-art methods in different data regimes.
\item We present a constrained decoding algorithm along with a set of key ingredients to guarantee the generation of valid and coherent data.
\item Our proposed solution is practical and can be expected to be applicable to other low-resource NLP tasks as it does not require any task-specific attributes.
\end{enumerate}{}

\section{Related Work}

\paragraph{Style Transfer}
Style transfer aims to adjust the stylistic characteristics of a sentence while preserving its original meaning. It has been widely studied in both supervised \citep{jhamtani-etal-2017-shakespearizing, niu-etal-2018-multi, rao-tetreault-2018-dear, wang-etal-2019-harnessing, wang-etal-2020-formality} and unsupervised manners \citep{DBLP:conf/nips/YangHDXB18, li-etal-2018-delete, prabhumoye-etal-2018-style, john-etal-2019-disentangled, dai-etal-2019-style, DBLP:conf/iclr/HeWNB20, krishna-etal-2020-reformulating, liu-etal-2021-learning}. Although style transfer is frequently utilized in many NLP tasks such as dialogue systems \cite{niu-bansal-2018-polite, zhu-etal-2021-neural}, sentiment transfer \citep{DBLP:conf/nips/ShenLBJ17, malmi-etal-2020-unsupervised}, and text debiasing \citep{ma-etal-2020-powertransformer, he-etal-2021-detect-perturb}, its application on data augmentation still remains understudied by the NLP community. To facilitate research in this direction, we study style transfer as data augmentation and propose a novel approach to explore transferring the style-related attributes of text to add more variations in the training data. 

\paragraph{Data Augmentation}
Data augmentation has been recently receiving increasing and widespread attention in the field of NER. The mainstream methods can be group into two categories: rule-based approaches \citep{bodapati-etal-2019-robustness, dai-adel-2020-analysis, lin-etal-2021-rockner, simoncini-spanakis-2021-seqattack}, and model-based approaches \citep{ding-etal-2020-daga, nie-etal-2020-named, zeng-etal-2020-counterfactual, chen-etal-2020-local, wenjing-etal-2021-improving, liu-etal-2021-mulda, wang-henao-2021-unsupervised, zhao-etal-2021-glara}. Although previous methods have shown promising results, they may not perform well in low-resource domains as the size and diversity of the training data are limited. Recent work have studied data augmentation by leveraging the data from high-resource domains. \citet{chen-etal-2021-data} investigated data transformation with implicit textual patterns while \citet{zhang-etal-2021-pdaln} explored replacing entities between domains with cross-domain anchor pairs. Motivated by their impressive results, we further study how to bridge the gap of data difference between domains and explore a better to leverage the data in high-resource domains for data augmentation.

\section{Problem Formulation and Preliminaries}
\label{sec: preliminary}
Considering a nonparallel NER dataset $\mathcal{D}$ consisting of source data $\mathcal{D}_{src}$ from a high-resource domain and target data $\mathcal{D}_{tgt}$ from a low-resource domain, training a model directly on $\mathcal{D}_{src}$ and evaluating on $\mathcal{D}_{tgt}$ is expected to give low prediction performance due to the stylistic differences (e.g., lexicons and syntax) between domains. In this work, we transform the data from a source domain to a target domain by changing text style and use the resulting transformed data to improve NER performance on $\mathcal{D}_{tgt}$. To this end, we assume access to a dataset $\mathcal{P}$ that contains pairs of parallel source and target sentences, respectively, to provide supervision signals for style transfer and a pre-trained generative model $G_{\theta}$ based on an encoder-decoder \citep{DBLP:conf/nips/VaswaniSPUJGKP17} architecture. Given an input sequence $x = \left \{x_{1}, x_{2}, ..., x_{N}\right \}$ of length $N$, the generator $G_{\theta}$ is pre-trained to maximize the conditional probability in an autoregressive manner:
\begin{equation*}
\left.\begin{aligned}
p_{\theta_{G}}(\hat{y}|x) = \prod_{i=1}^{M}p_{\theta_{G}}(\hat{y}_{i}|\hat{y}_{<i}, x)\\
\end{aligned}\right.
\end{equation*}

where $\hat{y} = \left \{\hat{y}_{1}, \hat{y}_{2}, ..., \hat{y}_{M}\right \}$ is the generated sentence of length $M$ that has the same meaning as $x$ but a different style.

\paragraph{Data Preparation}
Applying pre-trained generative models requires converting NER tokens and labels into a linearized format. In this work, we assume that the NER dataset follows the standard BIO labeling schema \citep{tjong-kim-sang-veenstra-1999-representing}. Previous work \citep{ding-etal-2020-daga, chen-etal-2021-data} has explored pre-pending each label to its corresponding token so that the model can capture the dependency and relationship between tokens and labels. However, we argue that this approach requires prohibitively large amounts of training data to adequately model the labeling schema. Recent work has also found that this format introduces too many hallucinated tokens and thus makes the learning problem significantly harder as the model needs to track token indices implicitly \citep{maynez-etal-2020-faithfulness, Raman2022TransformingST}. In the scenario where the number of annotated data is limited, the model may be confused with the labeling schema and tend to generate noisy samples. To mitigate this issue, we propose to linearize the data by only adding \texttt{<START\_ENTITY\_TYPE>} and \texttt{<END\_ENTITY\_TYPE>} special tokens to the beginning and the end of each entity span. For instance, the sample ``\texttt{A rainy day in {[}New{]}\textsubscript{B-LOC} {[}York{]}\textsubscript{I-LOC}}" will be converted to ``\texttt{A rainy day in <START\_LOC> New York <END\_LOC>}". Furthermore, we also prepend task prefixes to the beginning of the sentences to guide the direction of style transfer, where a prefix is a sequence of words for task description specifying the source and target styles of transfer (e.g., ``transfer source to target: ").

\begin{figure*}[ht]
\centering
\includegraphics[width=1.0\linewidth]{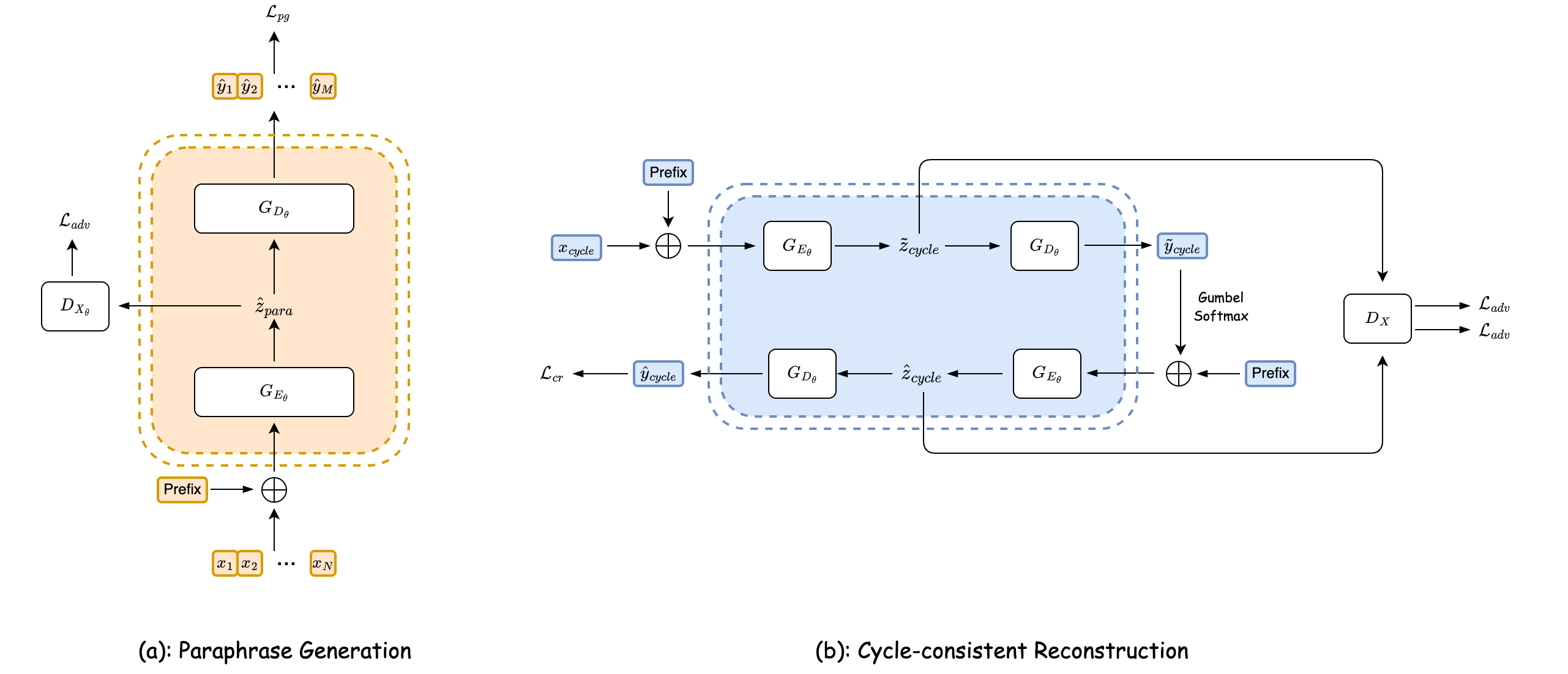}
\caption{The general architecture of our proposed method. Figure (a) shows the architecture for paraphrase generation, which aims to rephrase the sentence to a different style with supervision. Figure (b) shows the architecture for cycle-consistent reconstruction, which aims to transfer the sentence to a different style and then back to its original style. $G_{E_{\theta}}$ and $G_{D_{\theta}}$ refer to the encoder and decoder of the generator $G_{\theta}$ respectively while $D_{X_{\theta}}$ is the discriminator.}
\label{fig: model_architecture}
\end{figure*}

\section{Method}
\label{sec: method}
In this work, we explore style transfer as a data augmentation method to improve the performance of NER systems on low-resource domains. We propose an adversarial learning framework to generate a paraphrase of the source data in a high-resource domain whose style conforms to the target data in a low-resource domain. The proposed framework comprises two main components: (1) \textit{paraphrase generation}, which aims to rephrase the sentence to a different style with supervision, and (2) \textit{cycle-consistent reconstruction}, which aims to transfer the sentence to a different style and then back to its original style with no supervision. The overall architecture is shown in Figure \ref{fig: model_architecture}. Besides, we design a constrained decoding algorithm to guarantee the generation of valid samples and present a set of key ingredients to select high-quality generated sentences. 

\subsection{Adversarial Learning with PLMs}

\paragraph{Paraphrase Generation (PG)}
Recent work \citep{krishna-etal-2020-reformulating, qi-etal-2021-mind} has demonstrated the effectiveness of reformulating style transfer as a controlled paraphrase generation problem. Here, we follow the same idea to perform the style transfer task in a supervised manner by using a gold standard annotated corpus $\mathcal{P}$. Specifically, as shown in Figure \ref{fig: model_architecture}, given a sentence $x = \left \{x_{1}, x_{2}, ..., x_{N}\right \}$ of length $N$ concatenated with a task prefix that specifies the source and target styles of transfer, the generator $G_{\theta}$ encodes it into a sequence of latent representations $\hat{z}_{para}$ and then decodes these representations into its paraphrase $\hat{y} = \left \{\hat{y}_{1}, \hat{y}_{2}, ..., \hat{y}_{M}\right \}$ of length $M$ which has the same meaning as $x$ but a different style. The generator $G_{\theta}$ is trained with explicit supervision to maximize the log likelihood objective:
\begin{equation*}
\left.\begin{aligned}
& \mathcal{L}_{\text{pg}} = - \frac{1}{M} \sum_{i=1}^{M}  y_{i} \cdot  \text{log} \left( \hat{y}_{i} \right) \\
\end{aligned}\right.
\end{equation*}

\paragraph{Cycle-consistent Reconstruction (CR)}
Considering a nonparallel NER dataset consisting of data from $\mathcal{D}_{src}$ and $\mathcal{D}_{tgt}$ in the source and target styles, respectively, we aim to change the style of text as a way to augment data. Previous mechanism enables the generator $G_{\theta}$ to learn different mapping functions, i.e., $G_{\theta}(x_{src}) \to \hat{y}_{tgt}$ and $G_{\theta}(x_{tgt}) \to \hat{y}_{src}$. Intuitively, the learned mapping functions should be reverses of each other. The sentence transferred by one mapping function should be able to be transferred back to its original representation using the other mapping function. Such cycle-consistency can not only encourage content preservation between the input and output, but also reduce the search space of mapping functions \citep{DBLP:conf/nips/ShenLBJ17, prabhumoye-etal-2018-style}. To this end, as shown in Figure \ref{fig: model_architecture}, we first use the generator $G_{\theta}$ to generate the paraphrase $\tilde{y}_{cycle}$ of the input sentence $x_{cycle}$ concatenated with a prefix. As the gradients cannot be back-propagated through discrete tokens, we use Gumbel Softmax \citep{DBLP:conf/iclr/JangGP17} for $\tilde{y}_{cycle}$ as a continuous approximation to recursively sample the tokens from the probability distribution:
\begin{equation*}
\left.\begin{aligned}
& p_{\theta_{G}}(\hat{y}_{i}|\hat{y}_{<i}, x) = \frac{\text{exp}((\text{log}(\pi_{i}) + g_{i}) / \tau)}{\sum_{j=1}^{|V|}\text{exp}((\text{log}(\pi_{j}) + g_{j}) / \tau)} \\
\end{aligned}\right.
\end{equation*}
where $\pi$ are class probabilities for tokens and $|V|$ denotes the size of vocabulary. $g$ are i.i.d. samples drawn from Gumbel$(0, 1)$ distribution and $\tau$ is the temperature hyperparameter \citep{DBLP:journals/corr/HintonVD15}. $\tau \rightarrow 0$ approximates a one-hot representation while $\tau \rightarrow \infty$ approximates a uniform distribution. Then we concatenate the paraphrase $\tilde{y}_{cycle}$ with a different prefix as the input to the generator $G_{\theta}$ and let it transfer the paraphrase back to the original sentence $\hat{y}_{cycle}$. The training objective for cycle-consistent reconstruction is formulated as:
\begin{equation*}
\left.\begin{aligned}
\mathcal{L}_{\text{cr}} = 
& ~ \mathbb{E}_{x_{src} \sim X_{src}}[- \text{log} p_{\theta_{G}}(\tilde{y}_{tgt} | x_{src})] + \\ 
& ~ \mathbb{E}_{x_{tgt} \sim X_{tgt}}[- \text{log} p_{\theta_{G}}(\tilde{y}_{src} | x_{tgt})] \\
\end{aligned}\right.
\end{equation*}

Additionally, the generator $G_{\theta}$ is adversarially trained with a style discriminator $D_{\theta}$, which takes as the input the latent representations of either the input sentence or its paraphrase, and discriminates the input between the source and target styles:
\begin{equation*}
\left.\begin{aligned}
\mathcal{L}_{\text{adv}} = 
& ~ \mathbb{E}_{x_{src} \sim X_{src}}[- \text{log} D_{\theta}(G_{\theta}(x_{src}))] +  \\ 
& ~ \mathbb{E}_{x_{tgt} \sim X_{tgt}}[- \text{log} (1 - D_{\theta}(G_{\theta}(x_{tgt})))] \\
\end{aligned}\right.
\end{equation*}

\paragraph{Overall Training Objective}
The overall training objective can be formalized as:
\begin{equation*}
\left.\begin{aligned}
& \mathcal{L}(\theta_{G}, \theta_{D}) = \lambda_{\text{pg}} \mathcal{L}_{\text{pg}} + \lambda_{\text{cr}} \mathcal{L}_{\text{cr}} + \lambda_{\text{adv}} \mathcal{L}_{\text{adv}}\\
\end{aligned}\right.
\end{equation*}
where $\lambda_{\text{pg}}$, $\lambda_{\text{cr}}$, and $\lambda_{\text{adv}}$ reflect the relative importance of $\mathcal{L}_{\text{pg}}$, $\mathcal{L}_{\text{cr}}$, and $\mathcal{L}_{\text{adv}}$, respectively.

The training process begins with the paraphrase generation as the first stage: the generator $G_{\theta}$ is trained with the paraphrase generation objective while the discriminator $D_{\theta}$ is trained with the adversarial learning objective. In the second stage, both paraphrase generation and cycle-consistent reconstruction are involved: the cycle-consistent reconstruction objective is further incorporated to train the generator $G_{\theta}$.

\subsection{Data Transformation}
\begin{figure}[t]
\centering
\includegraphics[width=0.8\linewidth]{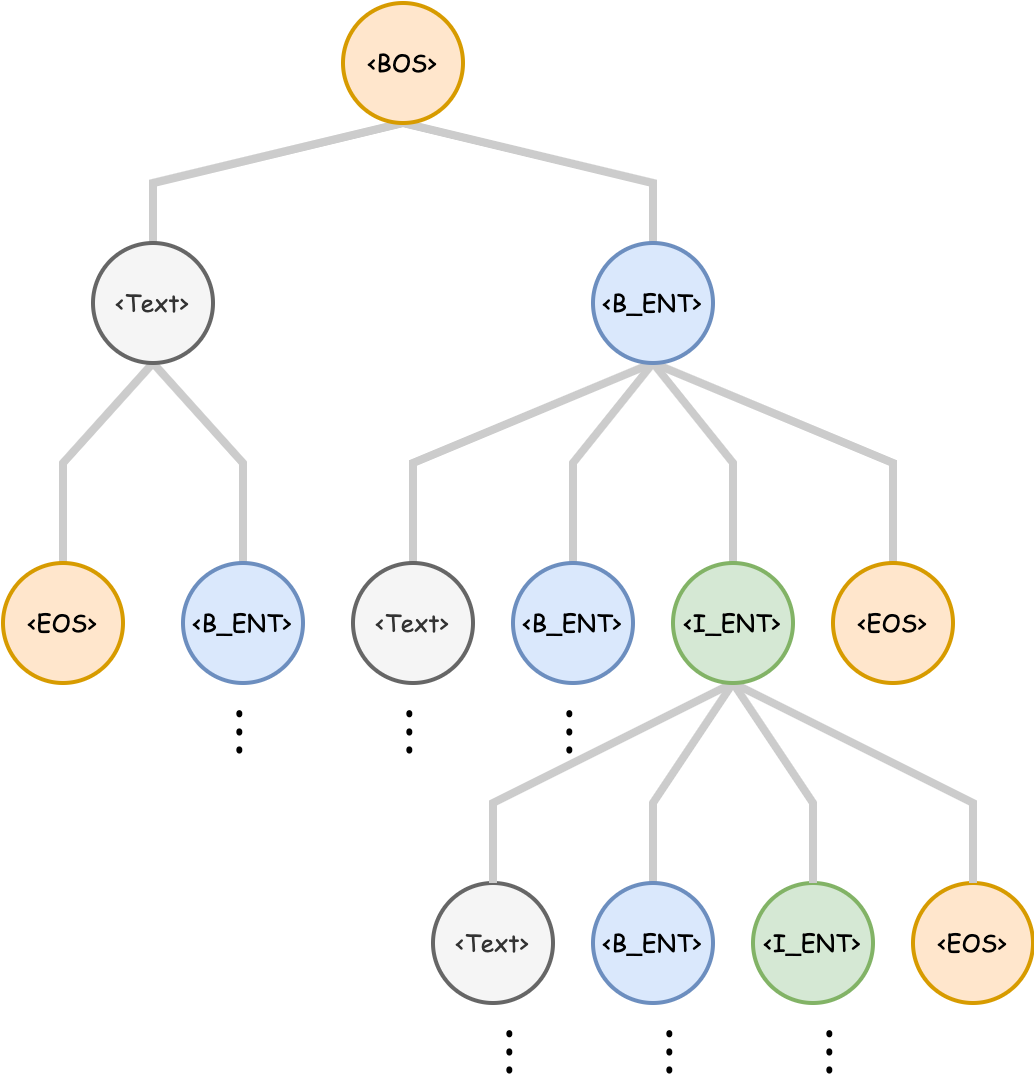}
\caption{
Prefix tree based constraints decoding algorithm for sentence generation. 
\texttt{<BOS>} and \texttt{<EOS>} represent the start and the end of each sentence. 
\texttt{<Text>} refers to the text spans while \texttt{<B\_ENT>} and \texttt{<I\_ENT>} refer to the start and inside of entity spans respectively. 
}
\label{fig: prefix_tree}
\end{figure}
\paragraph{Constrained Decoding}
When given the input sequence $x$ and the already generated tokens $\left \{y_{1}, y_{2}, ..., y_{i-1}\right \}$, a straightforward approach to generate the next word $y_{i}$ is to greedily select the word with the highest probability at the timestep $i$. However, this greedy decoding approach may result in the sub-optimal decision and cannot guarantee to generate valid NER samples (e.g, mismatch of entity types and incomplete sentences). 

To address these issues, we propose to apply a constrained decoding algorithm based on prefix trees \citep{DBLP:conf/iclr/CaoI0P21, lu-etal-2021-text2event} to control data transformation. Figure \ref{fig: prefix_tree} presents our proposed algorithm. Specifically, the decoding starts with a \texttt{<BOS>} token and ends with a \texttt{<EOS>} token. At each decoding step, we apply top-k \citep{fan-etal-2018-hierarchical} and top-p \citep{DBLP:conf/iclr/HoltzmanBDFC20} algorithms to navigate the search space. The prefix tree maintains two candidate vocabularies for text spans (i.e, \texttt{<Text>} node) and entity spans (i.e, \texttt{<B\_ENT>} and \texttt{<I\_ENT>} nodes), respectively. Based on the previous generated token, the constrained decoding dynamically prunes the vocabulary to lead the model to generate valid tokens. For example, if the previously generated token is \texttt{<Text>}, the model can only generate either \texttt{<EOS>} or \texttt{<B\_ENT>} as the next token. Otherwise, it makes the sample invalid and noisy. We also adapt a slightly larger temperature (i.e., $\tau$ in Gumbel Softmax) to smooth the probability distribution towards the tokens that most likely conform to the target style. 

\paragraph{Data Selection}
Even with a valid structure, the generated sentences may still remain unreliable as it may have a low quality due to degenerate repetition and incoherent gibberish \citep{DBLP:conf/iclr/HoltzmanBDFC20, DBLP:conf/iclr/WelleckKRDCW20}. To mitigate this issue, we further perform data selection with the following metrics: 
\begin{itemize}[topsep=3pt,itemsep=-0.5ex,partopsep=1ex,parsep=1ex]
	\item \textit{Consistency}: a confidence score from a pre-trained style classifier as the extent a generated sentence is in the target style.
	\item \textit{Adequacy}: a confidence score from a pre-trained NLU model on how much semantics is preserved in the generated sentence.
	\item \textit{Fluency}: a confidence score from a pre-trained NLU model indicating the fluency of the generated sentence.
	\item \textit{Diversity}: the edit distance between original sentences and the generated sentences at the character level.
\end{itemize}

For each sentence, we over-generate \textit{k}=10 candidates. We calculate the above metrics (see Appendix \ref{sec: score_calculation_in_data_selection} for more details) and assign a weighted score of these metrics to each candidate. Then we use the score to rank all candidates and select the best one for training NER systems.

\section{Experiments}
\begin{table*}[ht]
    \centering
    \renewcommand{\arraystretch}{0.8}
    \resizebox{\linewidth}{!}{
    \begin{tabular}{l|cccc|cccc|cccc}
    \toprule
    \multirow{2}{*}{\bf Method} & \multicolumn{4}{c}{\bf Few-shot Setting} & \multicolumn{4}{|c}{\bf Low-resource Setting} & \multicolumn{4}{|c}{\bf Full-set Setting} \\
    \cmidrule(lr){2-5}\cmidrule(lr){6-9}\cmidrule(lr){10-13}
    & \bf 1K & \bf 2K & \bf 3K & \bf 4K & \bf 1K & \bf 2K & \bf 3K & \bf 4K  & \bf 1K & \bf 2K & \bf 3K & \bf 4K \\
    \toprule
    
    \midrule
    \multicolumn{13}{l}{\textbf{\textit{BC $\rightarrow$ SM}}} \\
    \midrule
    Source (Baseline)                   & 33.13 & 35.07 & 36.37 & 36.52 & 33.13 & 35.07 & 36.37 & 36.52 & 33.13 & 35.07 & 36.37 & 36.52 \\
    \midrule
    ADA \citep{zhang-etal-2021-pdaln}   & 33.51 & 34.16 & 34.71 & 35.04 & 34.35 & 34.86 & 35.45 & 35.81 & 35.34 & 35.79 & 35.94 & 36.05 \\
    CDA \citep{chen-etal-2021-data}     & 35.21 & 37.14 & 39.48 & 39.59 & 36.59 & 40.75 & 42.23 & 43.26 & 45.04 & 49.81 & 52.06 & 53.81 \\
    Ours                                & 39.95 & 41.58 & 43.10 & 43.71 & 48.77 & 50.71 & 52.26 & 53.37 & 52.84 & 58.04 & 59.35 & 60.32 \\
    \midrule
    
    \multicolumn{13}{l}{\textbf{\textit{BN $\rightarrow$ SM}}} \\
    \midrule
    Source (Baseline)                   & 33.77 & 35.25 & 36.40 & 36.85 & 33.77 & 35.25 & 36.40 & 36.85 & 33.77 & 35.25 & 36.40 & 36.85 \\
    \midrule
    ADA \citep{zhang-etal-2021-pdaln}   & 31.81 & 34.31 & 35.35 & 36.08 & 34.39 & 35.66 & 36.10 & 36.48 & 35.57 & 36.40 & 36.97 & 37.39 \\
    CDA \citep{chen-etal-2021-data}      & 31.92 & 35.65 & 37.55 & 39.22 & 36.44 & 39.23 & 40.45 & 41.07 & 41.81 & 47.66 & 50.83 & 52.52 \\
    Ours                                & 36.47 & 40.04 & 40.72 & 41.38 & 44.87 & 47.57 & 50.29 & 51.52 & 53.56 & 56.81 & 59.01 & 59.40 \\
    \midrule
    
    \multicolumn{13}{l}{\textbf{\textit{MZ $\rightarrow$ SM}}} \\
    \midrule
    Source (Baseline)                   & 26.04 & 31.38 & 32.17 & 33.36 & 26.04 & 31.38 & 32.17 & 33.36 & 26.04 & 31.38 & 32.17 & 33.36 \\
    \midrule
    ADA \citep{zhang-etal-2021-pdaln}   & 28.92 & 32.83 & 34.19 & 35.07 & 30.35 & 33.66 & 35.10 & 35.78 & 32.97 & 35.13 & 35.87 & 35.99 \\
    CDA \citep{chen-etal-2021-data}     & 33.14 & 36.05 & 36.93 & 37.75 & 37.55 & 38.31 & 39.14 & 40.04 & 41.32 & 45.83 & 46.22 & 46.35 \\
    Ours                                & 34.71 & 38.10 & 40.71 & 41.97 & 47.88 & 52.11 & 52.82 & 53.24 & 51.04 & 55.43 & 57.12 & 58.00\\
    \midrule
    
    \multicolumn{13}{l}{\textbf{\textit{NW $\rightarrow$ SM}}} \\
    \midrule
    Source (Baseline) & 34.15 & 35.64 & 36.57 & 36.77 & 34.15 & 35.64 & 36.57 & 36.77 & 34.15 & 35.64 & 36.57 & 36.77 \\
    \midrule
    ADA \citep{zhang-etal-2021-pdaln}   & 34.05 & 34.70 & 35.47 & 35.94 & 34.78 & 35.84 & 36.18 & 36.45 & 35.29 & 36.77 & 37.50 & 37.62 \\
    CDA \citep{chen-etal-2021-data}     & 36.76 & 38.70 & 40.04 & 41.55 & 36.03 & 38.03 & 38.98 & 39.26 & 41.32 & 43.84 & 45.25 & 46.70 \\
    Ours                                & 41.66 & 45.97 & 48.73 & 49.40 & 47.58 & 50.65 & 52.00 & 52.86 & 54.07 & 56.43 & 57.14 & 57.29 \\
    \midrule
    
    \multicolumn{13}{l}{\textbf{\textit{WB $\rightarrow$ SM}}} \\
    \midrule
    Source (Baseline)                   &  8.23 & 12.93 & 15.08 & 15.96 &  8.23 & 12.93 & 15.08 & 15.96 &  8.23 & 12.93 & 15.08 & 15.96 \\
    \midrule
    ADA \citep{zhang-etal-2021-pdaln}   &  9.93 & 15.96 & 17.12 & 18.57 & 12.40 & 17.93 & 18.13 & 18.17 & 18.19 & 21.64 & 23.34 & 24.62 \\
    CDA \citep{chen-etal-2021-data}     & 15.95 & 17.06 & 18.91 & 19.89 & 15.04 & 21.12 & 24.64 & 26.45 & 24.61 & 27.75 & 29.64 & 30.07 \\
    Ours                                & 17.43 & 24.96 & 25.83 & 26.56 & 22.65 & 26.90 & 28.60 & 29.27 & 27.59 & 37.27 & 38.90 & 39.70 \\
    \midrule
    
    \midrule
    \bf Target (SM) & - & - & - & - & - & - & - & - & 68.42 & 73.49 & 75.36 & 76.63 \\
    \midrule
    
    \toprule
    \end{tabular}
    }
    \caption{Performance comparison of different data augmentation methods with same amount of pseudo data for training. Scores are calculated with the micro F1 metric.}
    \label{tab: data_quality}
\end{table*}
In this section, we present the experimental setup and results. We extensively evaluate our proposed method on five different domain pairs and compare our proposed method with state-of-the-art systems on data augmentation for NER in cross-domain scenarios.

\subsection{Experimental Setup}

\paragraph{Datasets}
We focus exclusively on formality style transfer which aims to transfer formal sentences to informal sentences. We use the parallel style transfer data from the GYAFC \footnote{\url{https://github.com/raosudha89/GYAFC-corpus}} \citep{rao-tetreault-2018-dear} as $\mathcal{P}$. This corpus contains pairs of formal and informal sentences collected from Yahoo Answers. For the nonparallel NER data $\mathcal{D}$, we use a subset of OntoNotes 5.0 \footnote{\url{https://catalog.ldc.upenn.edu/LDC2013T19}} corpus as source data $\mathcal{D}_{src}$ and Temporal Twitter Corpus \footnote{\url{https://github.com/shrutirij/temporal-twitter-corpus}} \citep{rijhwani-preotiuc-pietro-2020-temporally} as target data $\mathcal{D}_{tgt}$. Here, we only consider the English datasets. The source data involves five different domains in the formal style: broadcast
conversation (\texttt{BC}), broadcast news (\texttt{BN}), magazine (\texttt{MZ}), newswire (\texttt{NW}), and web data (\texttt{WB}) while the target data involves only social media (\texttt{SM}) domain in the informal style. The source and target data are nonparallel and we consider a total of 18 different entity types (e.g, \textit{PERSON} and \textit{LOCATION}). The data statistics and a list of entity types are shown in Appendix \ref{sec: data_statistics}. The details of data preprocessing and filtering are described in Appendix \ref{sec: data_preprocessing_and_filtering}.

\begin{table*}[ht]
    \centering
    \renewcommand{\arraystretch}{1.2}
    \resizebox{\linewidth}{!}{
    \begin{tabular}{ccccc}
    \toprule
    \textbf{BC $\rightarrow$ SM} & \textbf{BN $\rightarrow$ SM} & \textbf{MZ $\rightarrow$ SM} & \textbf{NW $\rightarrow$ SM} & \textbf{WB $\rightarrow$ SM} \\
    \toprule
    \multicolumn{5}{c}{\includegraphics[scale=0.35]{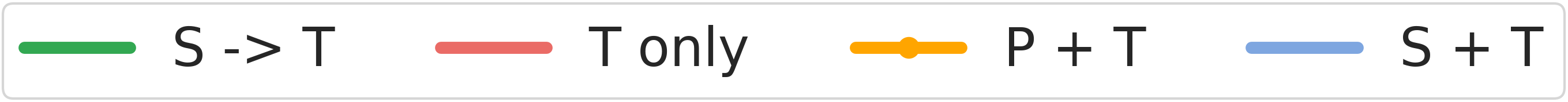}} \\
    \includegraphics[scale=0.35]{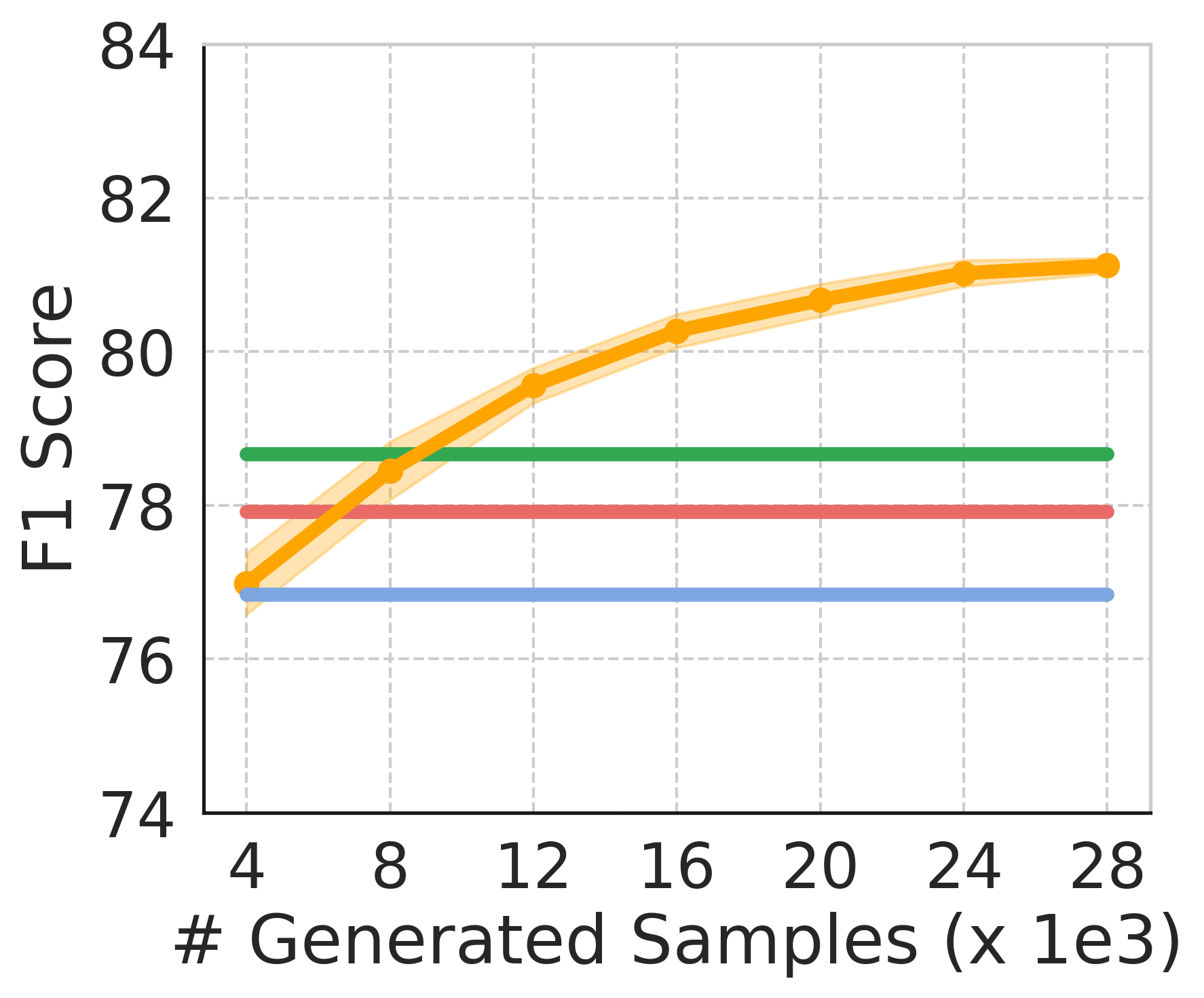} & \includegraphics[scale=0.35]{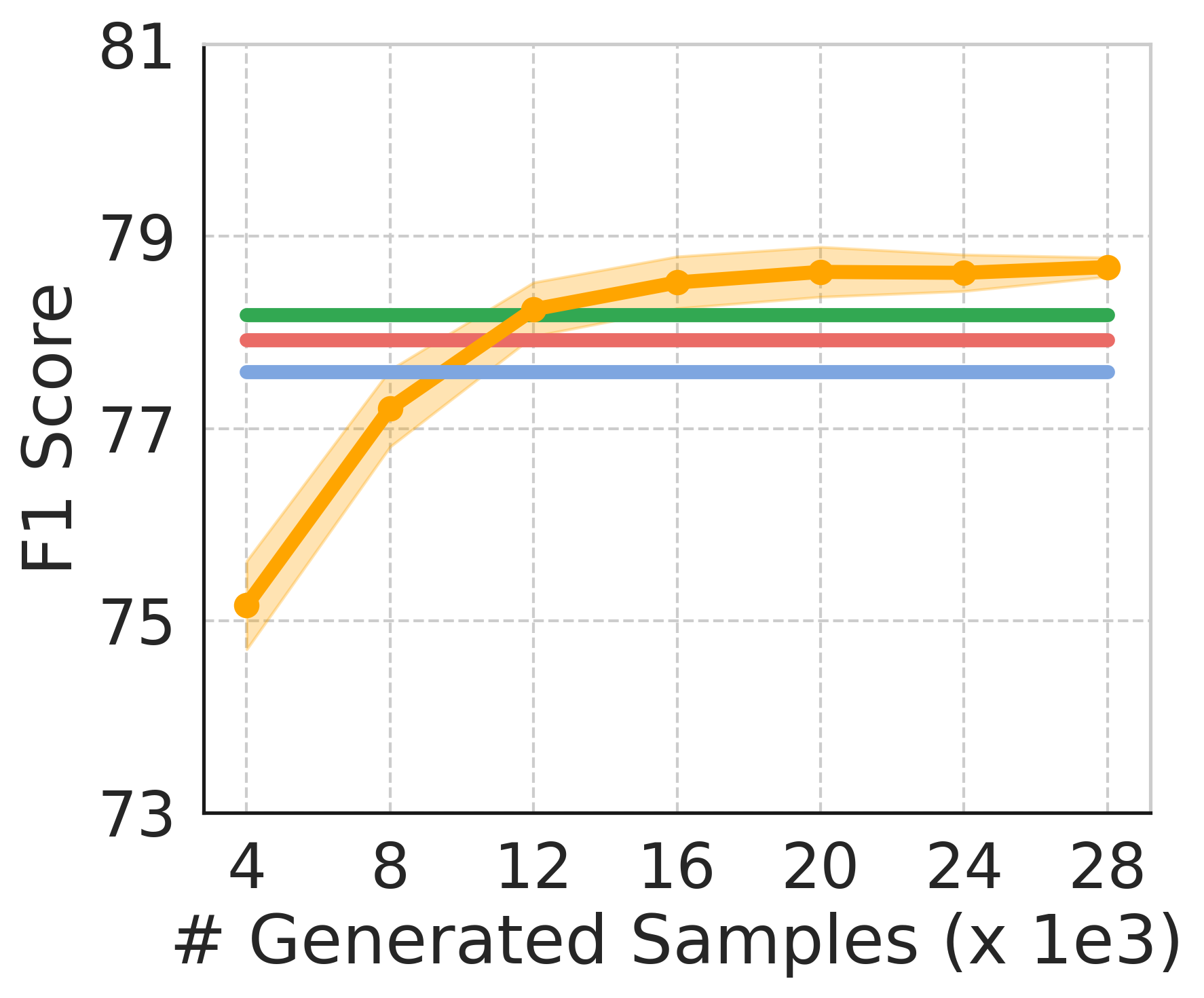} & \includegraphics[scale=0.35]{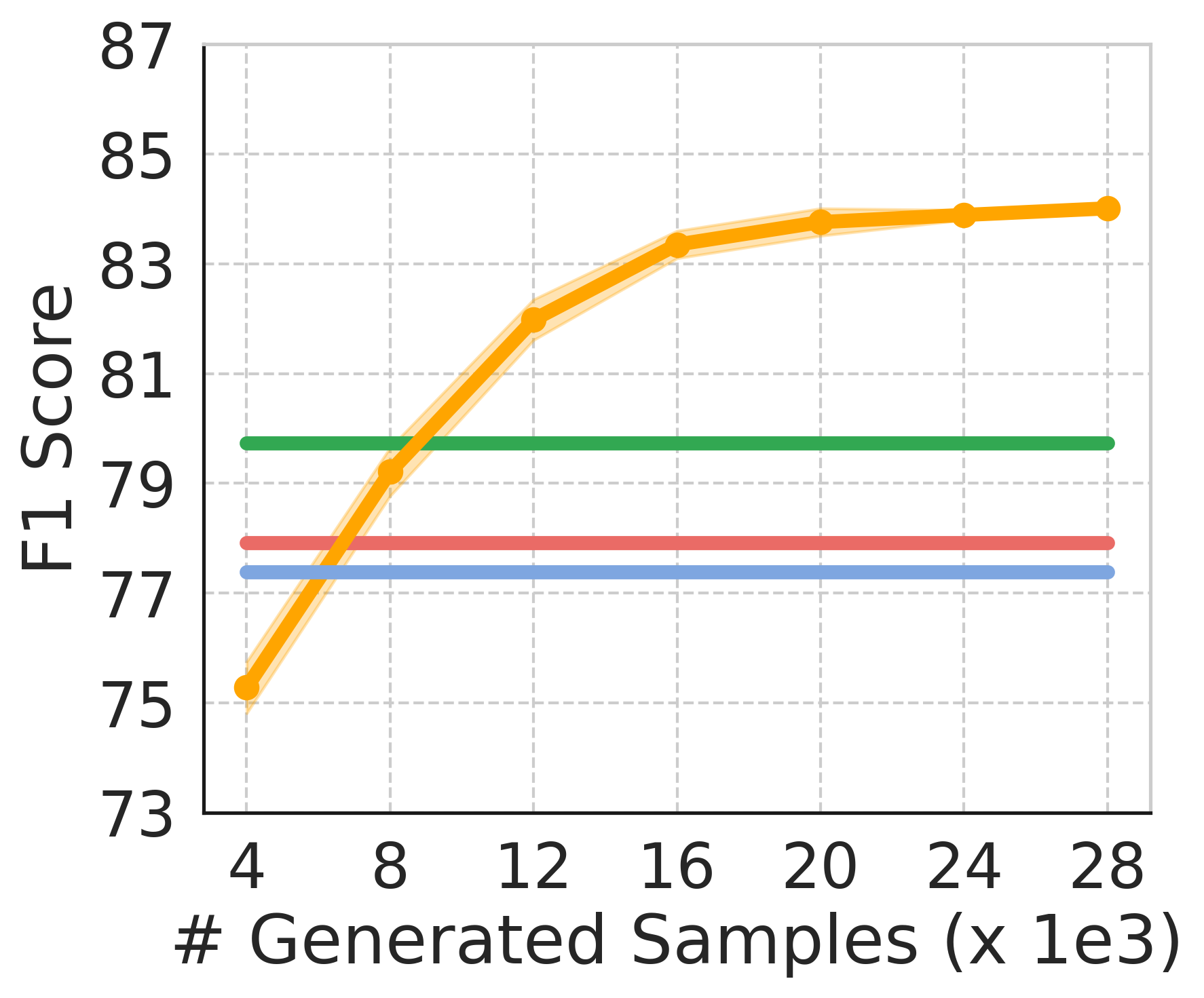} & \includegraphics[scale=0.35]{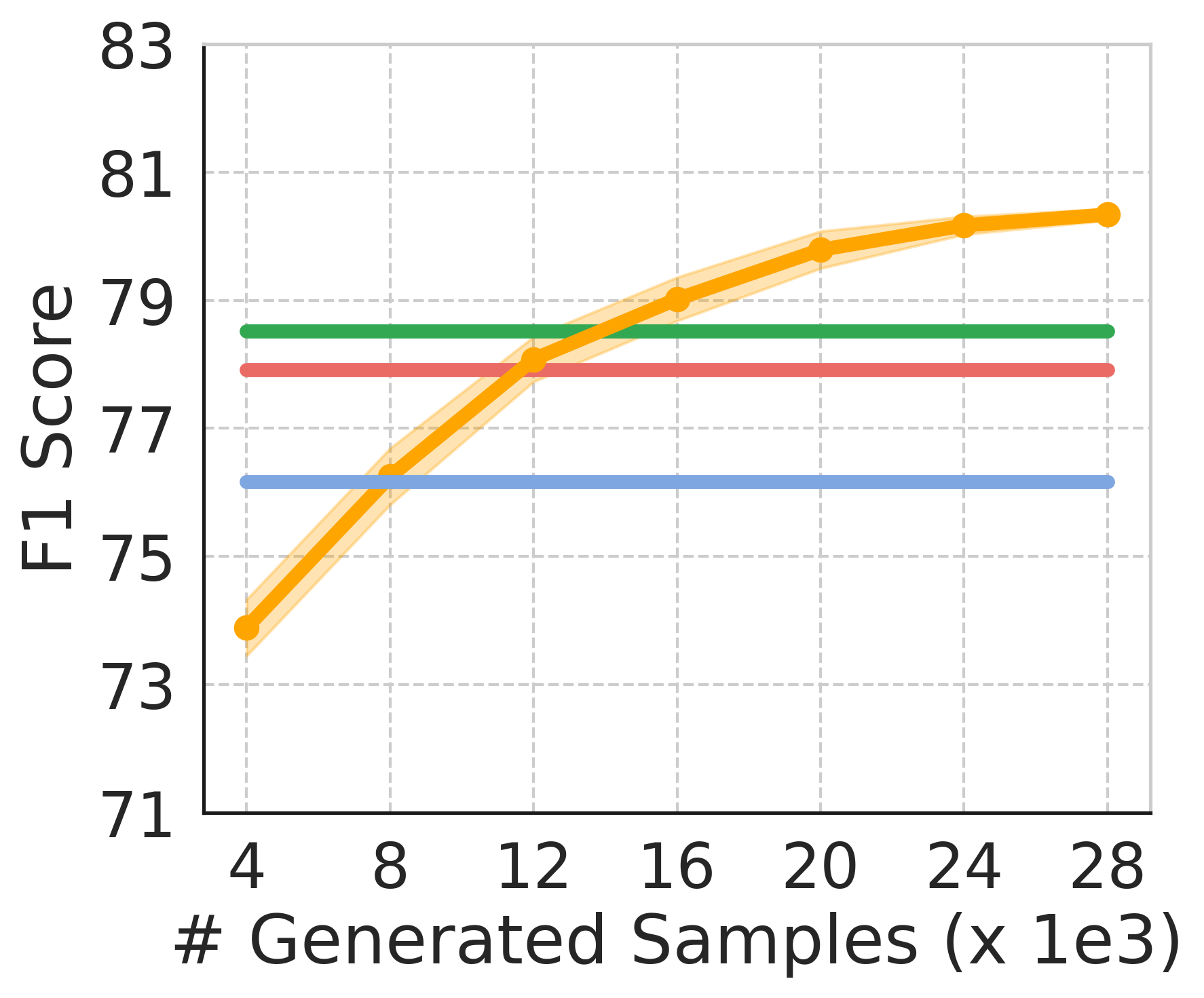} & \includegraphics[scale=0.35]{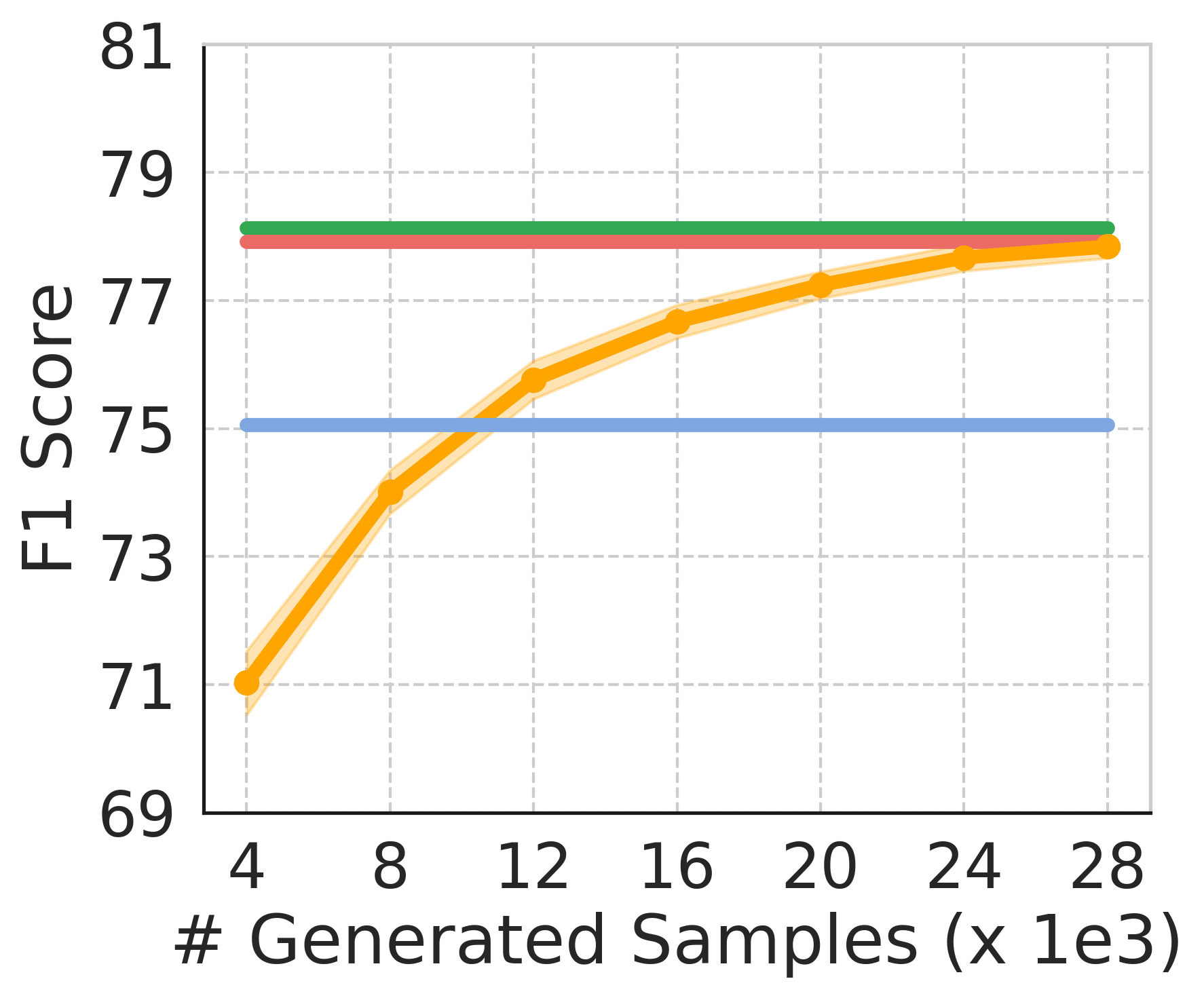} \\
    \toprule
    \end{tabular}
    }
    \caption{Performance comparison of different data augmentation methods with large amount of pseudo data for training in \textsc{full-set} setting: (1) \textbf{S + T}: fine-tune on source and target data together, (2) \textbf{T only}: fine-tune on only target data, (3) \textbf{S $\rightarrow$ T}: fine-tune on source data first and then target data, and (4) \textbf{P + T}, fine-tune on pseudo data and target data together while the number of samples from pseudo data is different in different steps. Scores are calculated with the micro F1 metric.}
    \label{tab: data_reliability}
\end{table*}

\paragraph{Base Models}
For style transfer, we use a pre-trained $\text{T5}_{\text{base}}$ model \citep{2020t5} to initialize both the generator $G_{\theta}$ and discriminator $D_{\theta}$. For NER, we use a sequence labeling framework consisting of a pre-trained $\text{BERT}_{\text{base}}$ model \citep{devlin-etal-2019-bert} as the text encoder and a linear layer as the classifier to assign labels for each token. We use Huggingface Transformers library \citep{wolf-etal-2020-transformers} to implement all models. The details of hyper-parameters and fine-tuning are described in Appendix \ref{sec: hyperparameters_and_finetuning}. 

\paragraph{Data Regimes}
To better understand the effectiveness of our proposed method, we undertake experiments in three different data regimes by varying the amount of training data in the target domain, namely \textsc{few-shot}, \textsc{low-resource}, and \textsc{full-set} scenarios. For all scenarios, we assume full access to $\mathcal{P}$ and $\mathcal{D}_{src}$ but different access to $\mathcal{D}_{tgt}$. In the \textsc{few-shot} scenario, we adopt a $N$-way $K \sim 2K$-shot setting following \citet{ding-etal-2021-nerd}. We randomly select samples from $\mathcal{D}_{tgt}$ and ensure that each entity class contains $K \sim 2K$ examples. The \textit{K} is set to 10 in our experiments and thus we will have 10 $\sim$ 20 samples from the target data. In the \textsc{low-resource} scenario, we simulate low-resource settings by randomly selecting 1024 samples from $\mathcal{D}_{tgt}$. For the \textsc{full-set} scenario, we assume a full access to $\mathcal{D}_{tgt}$, i.e., use all of samples from the target data. 

\paragraph{Compared Methods}
We investigate the following methods on data augmentation for NER in cross-domain scenarios for comparison: (1) \textbf{Adaptive Data Augmentation (ADA)} \citep{zhang-etal-2021-pdaln} which proposes to augment sentences by replacing the entity in the source data with the entity in the target data that belongs to same entity class, and (2) \textbf{Cross-domain Data Augmentation (CDA)} \citep{chen-etal-2021-data} which studies to augment sentences by transforming data representations through an aligned feature space between the source and target data. We apply each method on source data to obtain the same amount of generated pseudo data. Each sample in the pseudo data corresponds to a sample in the source data. To make a fair comparison, we use the same base model (i.e., $\text{BERT}_{\text{base}} + \text{Linear}$) but different training data generated from each method. The validation and test data are from the target domain. We do five runs for all experiments and report the average micro F1 score as the evaluation metric.

\subsection{Main Results}
\paragraph{Using Same Amount of Pseudo Data}
Here, we randomly select 1K, 2K, 3K, and 4K samples generated by each method as the training data to fine-tune the model. The baseline is established by fine-tuning the model on the same amount of data from the source domain. The validation and test data are from the target domain. Table \ref{tab: data_quality} presents the performance comparison of different data augmentation methods with the same amount of pseudo data. Overall, our proposed method significantly outperforms the previous best method. On average, the F1 score increases by 4.7\%, 10.1\%, and 9.3\% in \textsc{few-shot}, \textsc{low-resource}, and \textsc{full-set} settings, respectively. Regarding the effect of data augmentation in cross-domain settings, we find that simply replacing the entities may decrease the model performance, especially when we only have limited amounts of target data. Although this method can comparatively address the word discrepancy by sharing entities across domains, it potentially increases the overlap between the training and test data, and thus reduces the model's generalization ability. Additionally, learning text differences between domains from only nonparallel samples suffers from the problem of data scarcity and can result in the generation of invalid and/or low-quality data. In contrast, our proposed method is more stable and consistently outperforms the baseline in different settings. We attribute this improvement to the fact that the proposed method can significantly increase the diversity from the perspective of words and entities while barely bringing semantic changes to the original text.

\paragraph{Using Large Amount of Pseudo Data}
Theoretically, we could generate an infinite amount of pseudo data for training. Thus, we undertake experiments using more pseudo data combined with target data for training. Here, we make comparison with three different method to support the effectiveness of our proposed method: (1) \textbf{S + T}: fine-tune on source and target data together, (2) \textbf{T only}: fine-tune on only target data, and (3) \textbf{S $\rightarrow$ T}: fine-tune on source data first and then target data. For our proposed method \textbf{P + T}, we gradually increase the number of generated samples combined with target data as for training. We present the results in Figure \ref{tab: data_reliability}. Notably, combining source data directly with target data could hurt the model performance. One possible explanation for such poor performance is that the distribution (e.g., lexicons and syntax) of source and target data can be very different, which may encourage the model to learn irrelevant patterns and thus lead to under-fitting. We also notice a similar phenomenon while only using a few amounts of samples combined with target samples. We argue that, in such cases, the model can have an inductive bias towards memorization instead of generalization, and thus perform poorly on test data. Additionally, fine-tuning on the source data first and then target data \textbf{S $\rightarrow$ T} can achieve better results than simply fine-tuning on source and target data together \textbf{S + T} or only target data \textbf{T only}. Nevertheless, with more and more generated samples for training, our proposed method can significantly boost the model performance comparing against other methods in four domain pairs (\texttt{BC}, \texttt{BN}, \texttt{MZ}, and \texttt{NW}) while remains very competitive in \texttt{WB}.

\subsection{Analysis}
\paragraph{Ablation Study}
In Table \ref{tab: ablation_study}, we present ablation studies in the \textsc{full-set} setting. For each domain pair, we generate the training data by randomly transferring 1K samples from the source domain while the validation and test data are from the target domain \texttt{SM}. We consider a series of ablation studies: (1) no cycle-consistent reconstruction, which indicates that we train the model with only the style transfer datasets, (2) no style discriminator, (3) no constrained decoding (i.e., no guarantee on the generation of valid sentences, and (4) no data selection. From Table \ref{tab: ablation_study}, we can see that cycle-consistent reconstruction is critical for our proposed method to transfer the knowledge between domains. Without this component, the F1 score decreases by 39.15\% on average. Additionally, constrained decoding also plays an important role in avoiding label mistakes, which can significantly hurt model performance. Moreover, the style discriminator is effective to enable the model to find generalized patterns while data selection can further boost the model performance. Overall, the ablation results demonstrate that each strategy in our proposed method is crucial in achieving promising results.

\newcommand*{\tab}{\hspace*{0.3cm}}

\begin{table}[t]
    \centering
    \renewcommand{\arraystretch}{1.2}
    \resizebox{\linewidth}{!}{
        \begin{tabular}{lccccc}
            \toprule
            \bf Method & \bf BC & \bf BN & \bf MZ & \bf NW & \bf WB \\
            \toprule
            Ours                        & 52.84 & 53.56 & 51.04 & 54.07 & 27.59 \\
            \toprule
            \tab - CR                   & 15.30 & 13.82 &  7.11 &  4.15 &  2.98 \\
            \tab - style discriminator  & 35.79 & 23.87 & 29.39 & 19.58 & 11.47 \\
            \tab - constrained decoding & 18.91 &  7.56 & 12.66 & 15.81 & 13.73 \\
            \tab - data selection       & 46.48 & 52.31 & 48.12 & 49.30 & 20.81 \\
            \toprule
        \end{tabular}
    }
    \caption{Ablation study with different domain as source data and \texttt{SM} as target data. CR refers to cycle-consistent reconstruction. Scores are calculated with the micro F1 metric.}
    \label{tab: ablation_study}
\end{table}

\paragraph{Case Study}
Table \ref{tab: case_study} shows some hand-picked examples of formal source sentences and their corresponding informal target sentences generated from our proposed method. We observe that the generated sentences are embedded with some target domain characteristics (e.g., misspellings, grammar errors, language variations, and emojis) which significantly enhances entity context yet remains semantically related and coherent. This indicates that our proposed method can learn and transfer the text styles. Besides, we find that some entities in the original sentences can be replaced with not only those of the same entity type but also those of a different entity type. We also noticed that the proposed method tends to generate short sentences, imitating the target data to make the context ambiguous. However, we also note that the model may ignore the entity spans in the original sentences, which has a negative impact on the diversity of entities if the data is very limited. Besides, our model cannot deal well with the mismatch of labeling schema (i.e., the same entity span is labeled into different entity types in the source and target data) and has a limited ability to recognize rare entities. In the future, we plan to continue exploring approaches to address these issues.

\section{Conclusion}
In this paper, we propose an effective approach to employ style transfer as a data augmentation method. Specifically, we present an adversarial learning framework to bridge the gap between different text styles and transfer the entity knowledge from high-resource domains to low-resource domains. Additionally, to guarantee the generation of valid and coherent data, we design a constrained decoding algorithm along with a set of key ingredients for data generation and selection. We undertake experiments on five different domain pairs. The experimental results show that our proposed method can effectively transfer the data across domains to increase the size and diversity of training data for low-resource NER tasks. For the future work, we plan to explore style transfer as data augmentation in cross-lingual settings (i.e., transfer entity knowledge across languages instead of just domains). Additionally, since our approach is based on pre-trained language models, it would be interesting to explore leveraging pre-trained knowledge for data augmentation.

\section*{Limitations}
Based on our studies, we find the following main limitations: (1) \textit{mismatch of annotation schema}: we observe that the annotation schema between some NER datasets conflict with each other. The same entity span can be labeled into different entity types. For example, ``America" is an instance of \textit{GPE} in the OntoNotes 5.0 dataset but \textit{LOCATION} in the WNUT17 dataset. This phenomenon introduces noise and make it difficult for models to understand entity types and learn transformations. (2) \textit{mismatch of labeling schema}: the labeling schema in different NER datasets can be very different. For instance, OntoNotes 5.0 dataset contains 18 coarse-grained entity types while FEW-NERD contains 9 coarse-grained and 66 fine-grained entity types. Using such datasets as source and target data may not lead to a significant improvement gains. We hope our findings can inform potential avenues of improvement on data augmentation for NER and inspire the further work in this research direction.

\section*{Acknowledgements}
This work was partially supported by the National Science Foundation (NSF) under grant \#1910192. We would like to thank the anonymous reviewers for their valuable suggestions and the members from the RiTUAL lab at the University of Houston for their valuable feedback. 

\bibliography{anthology,custom}

\begin{thebibliography}{63}
\expandafter\ifx\csname natexlab\endcsname\relax\def\natexlab#1{#1}\fi

\bibitem[{Aghajanyan et~al.(2021)Aghajanyan, Shrivastava, Gupta, Goyal,
  Zettlemoyer, and Gupta}]{DBLP:conf/iclr/AghajanyanSGGZG21}
Armen Aghajanyan, Akshat Shrivastava, Anchit Gupta, Naman Goyal, Luke
  Zettlemoyer, and Sonal Gupta. 2021.
\newblock \href {https://openreview.net/forum?id=OQ08SN70M1V} {Better
  fine-tuning by reducing representational collapse}.
\newblock In \emph{9th International Conference on Learning Representations,
  {ICLR} 2021, Virtual Event, Austria, May 3-7, 2021}. OpenReview.net.

\bibitem[{Bodapati et~al.(2019)Bodapati, Yun, and
  Al-Onaizan}]{bodapati-etal-2019-robustness}
Sravan Bodapati, Hyokun Yun, and Yaser Al-Onaizan. 2019.
\newblock \href {https://doi.org/10.18653/v1/D19-5531} {Robustness to
  capitalization errors in named entity recognition}.
\newblock In \emph{Proceedings of the 5th Workshop on Noisy User-generated Text
  (W-NUT 2019)}, pages 237--242, Hong Kong, China. Association for
  Computational Linguistics.

\bibitem[{Cao et~al.(2021)Cao, Izacard, Riedel, and
  Petroni}]{DBLP:conf/iclr/CaoI0P21}
Nicola~De Cao, Gautier Izacard, Sebastian Riedel, and Fabio Petroni. 2021.
\newblock \href {https://openreview.net/forum?id=5k8F6UU39V} {Autoregressive
  entity retrieval}.
\newblock In \emph{9th International Conference on Learning Representations,
  {ICLR} 2021, Virtual Event, Austria, May 3-7, 2021}. OpenReview.net.

\bibitem[{Chen et~al.(2020)Chen, Wang, Tian, Yang, and
  Yang}]{chen-etal-2020-local}
Jiaao Chen, Zhenghui Wang, Ran Tian, Zichao Yang, and Diyi Yang. 2020.
\newblock \href {https://doi.org/10.18653/v1/2020.emnlp-main.95} {Local
  additivity based data augmentation for semi-supervised {NER}}.
\newblock In \emph{Proceedings of the 2020 Conference on Empirical Methods in
  Natural Language Processing (EMNLP)}, pages 1241--1251, Online. Association
  for Computational Linguistics.

\bibitem[{Chen et~al.(2021)Chen, Aguilar, Neves, and
  Solorio}]{chen-etal-2021-data}
Shuguang Chen, Gustavo Aguilar, Leonardo Neves, and Thamar Solorio. 2021.
\newblock \href {https://doi.org/10.18653/v1/2021.emnlp-main.434} {Data
  augmentation for cross-domain named entity recognition}.
\newblock In \emph{Proceedings of the 2021 Conference on Empirical Methods in
  Natural Language Processing}, pages 5346--5356, Online and Punta Cana,
  Dominican Republic. Association for Computational Linguistics.

\bibitem[{Dai et~al.(2019)Dai, Liang, Qiu, and Huang}]{dai-etal-2019-style}
Ning Dai, Jianze Liang, Xipeng Qiu, and Xuanjing Huang. 2019.
\newblock \href {https://doi.org/10.18653/v1/P19-1601} {Style transformer:
  Unpaired text style transfer without disentangled latent representation}.
\newblock In \emph{Proceedings of the 57th Annual Meeting of the Association
  for Computational Linguistics}, pages 5997--6007, Florence, Italy.
  Association for Computational Linguistics.

\bibitem[{Dai and Adel(2020)}]{dai-adel-2020-analysis}
Xiang Dai and Heike Adel. 2020.
\newblock \href {https://doi.org/10.18653/v1/2020.coling-main.343} {An analysis
  of simple data augmentation for named entity recognition}.
\newblock In \emph{Proceedings of the 28th International Conference on
  Computational Linguistics}, pages 3861--3867, Barcelona, Spain (Online).
  International Committee on Computational Linguistics.

\bibitem[{Dehouck and
  G{\'o}mez-Rodr{\'\i}guez(2020)}]{dehouck-gomez-rodriguez-2020-data}
Mathieu Dehouck and Carlos G{\'o}mez-Rodr{\'\i}guez. 2020.
\newblock \href {https://doi.org/10.18653/v1/2020.coling-main.339} {Data
  augmentation via subtree swapping for dependency parsing of low-resource
  languages}.
\newblock In \emph{Proceedings of the 28th International Conference on
  Computational Linguistics}, pages 3818--3830, Barcelona, Spain (Online).
  International Committee on Computational Linguistics.

\bibitem[{Devlin et~al.(2019)Devlin, Chang, Lee, and
  Toutanova}]{devlin-etal-2019-bert}
Jacob Devlin, Ming-Wei Chang, Kenton Lee, and Kristina Toutanova. 2019.
\newblock \href {https://doi.org/10.18653/v1/N19-1423} {{BERT}: Pre-training of
  deep bidirectional transformers for language understanding}.
\newblock In \emph{Proceedings of the 2019 Conference of the North {A}merican
  Chapter of the Association for Computational Linguistics: Human Language
  Technologies, Volume 1 (Long and Short Papers)}, pages 4171--4186,
  Minneapolis, Minnesota. Association for Computational Linguistics.

\bibitem[{Ding et~al.(2020)Ding, Liu, Bing, Kruengkrai, Nguyen, Joty, Si, and
  Miao}]{ding-etal-2020-daga}
Bosheng Ding, Linlin Liu, Lidong Bing, Canasai Kruengkrai, Thien~Hai Nguyen,
  Shafiq Joty, Luo Si, and Chunyan Miao. 2020.
\newblock \href {https://doi.org/10.18653/v1/2020.emnlp-main.488} {{DAGA}: Data
  augmentation with a generation approach for low-resource tagging tasks}.
\newblock In \emph{Proceedings of the 2020 Conference on Empirical Methods in
  Natural Language Processing (EMNLP)}, pages 6045--6057, Online. Association
  for Computational Linguistics.

\bibitem[{Ding et~al.(2021)Ding, Xu, Chen, Wang, Han, Xie, Zheng, and
  Liu}]{ding-etal-2021-nerd}
Ning Ding, Guangwei Xu, Yulin Chen, Xiaobin Wang, Xu~Han, Pengjun Xie, Haitao
  Zheng, and Zhiyuan Liu. 2021.
\newblock \href {https://doi.org/10.18653/v1/2021.acl-long.248} {Few-{NERD}: A
  few-shot named entity recognition dataset}.
\newblock In \emph{Proceedings of the 59th Annual Meeting of the Association
  for Computational Linguistics and the 11th International Joint Conference on
  Natural Language Processing (Volume 1: Long Papers)}, pages 3198--3213,
  Online. Association for Computational Linguistics.

\bibitem[{Fan et~al.(2018)Fan, Lewis, and Dauphin}]{fan-etal-2018-hierarchical}
Angela Fan, Mike Lewis, and Yann Dauphin. 2018.
\newblock \href {https://doi.org/10.18653/v1/P18-1082} {Hierarchical neural
  story generation}.
\newblock In \emph{Proceedings of the 56th Annual Meeting of the Association
  for Computational Linguistics (Volume 1: Long Papers)}, pages 889--898,
  Melbourne, Australia. Association for Computational Linguistics.

\bibitem[{Gururangan et~al.(2020)Gururangan, Marasovi{\'c}, Swayamdipta, Lo,
  Beltagy, Downey, and Smith}]{gururangan-etal-2020-dont}
Suchin Gururangan, Ana Marasovi{\'c}, Swabha Swayamdipta, Kyle Lo, Iz~Beltagy,
  Doug Downey, and Noah~A. Smith. 2020.
\newblock \href {https://doi.org/10.18653/v1/2020.acl-main.740} {Don{'}t stop
  pretraining: Adapt language models to domains and tasks}.
\newblock In \emph{Proceedings of the 58th Annual Meeting of the Association
  for Computational Linguistics}, pages 8342--8360, Online. Association for
  Computational Linguistics.

\bibitem[{He et~al.(2020)He, Wang, Neubig, and
  Berg{-}Kirkpatrick}]{DBLP:conf/iclr/HeWNB20}
Junxian He, Xinyi Wang, Graham Neubig, and Taylor Berg{-}Kirkpatrick. 2020.
\newblock \href {https://openreview.net/forum?id=HJlA0C4tPS} {A probabilistic
  formulation of unsupervised text style transfer}.
\newblock In \emph{8th International Conference on Learning Representations,
  {ICLR} 2020, Addis Ababa, Ethiopia, April 26-30, 2020}. OpenReview.net.

\bibitem[{He et~al.(2021)He, Majumder, and
  McAuley}]{he-etal-2021-detect-perturb}
Zexue He, Bodhisattwa~Prasad Majumder, and Julian McAuley. 2021.
\newblock \href {https://doi.org/10.18653/v1/2021.findings-emnlp.352} {Detect
  and perturb: Neutral rewriting of biased and sensitive text via
  gradient-based decoding}.
\newblock In \emph{Findings of the Association for Computational Linguistics:
  EMNLP 2021}, pages 4173--4181, Punta Cana, Dominican Republic. Association
  for Computational Linguistics.

\bibitem[{Hinton et~al.(2015)Hinton, Vinyals, and
  Dean}]{DBLP:journals/corr/HintonVD15}
Geoffrey~E. Hinton, Oriol Vinyals, and Jeffrey Dean. 2015.
\newblock \href {http://arxiv.org/abs/1503.02531} {Distilling the knowledge in
  a neural network}.
\newblock \emph{CoRR}, abs/1503.02531.

\bibitem[{Holtzman et~al.(2020)Holtzman, Buys, Du, Forbes, and
  Choi}]{DBLP:conf/iclr/HoltzmanBDFC20}
Ari Holtzman, Jan Buys, Li~Du, Maxwell Forbes, and Yejin Choi. 2020.
\newblock \href {https://openreview.net/forum?id=rygGQyrFvH} {The curious case
  of neural text degeneration}.
\newblock In \emph{8th International Conference on Learning Representations,
  {ICLR} 2020, Addis Ababa, Ethiopia, April 26-30, 2020}. OpenReview.net.

\bibitem[{Jang et~al.(2017)Jang, Gu, and Poole}]{DBLP:conf/iclr/JangGP17}
Eric Jang, Shixiang Gu, and Ben Poole. 2017.
\newblock \href {https://openreview.net/forum?id=rkE3y85ee} {Categorical
  reparameterization with gumbel-softmax}.
\newblock In \emph{5th International Conference on Learning Representations,
  {ICLR} 2017, Toulon, France, April 24-26, 2017, Conference Track
  Proceedings}. OpenReview.net.

\bibitem[{Jhamtani et~al.(2017)Jhamtani, Gangal, Hovy, and
  Nyberg}]{jhamtani-etal-2017-shakespearizing}
Harsh Jhamtani, Varun Gangal, Eduard Hovy, and Eric Nyberg. 2017.
\newblock \href {https://doi.org/10.18653/v1/W17-4902} {Shakespearizing modern
  language using copy-enriched sequence to sequence models}.
\newblock In \emph{Proceedings of the Workshop on Stylistic Variation}, pages
  10--19, Copenhagen, Denmark. Association for Computational Linguistics.

\bibitem[{Jiang et~al.(2020)Jiang, He, Chen, Liu, Gao, and
  Zhao}]{jiang-etal-2020-smart}
Haoming Jiang, Pengcheng He, Weizhu Chen, Xiaodong Liu, Jianfeng Gao, and Tuo
  Zhao. 2020.
\newblock \href {https://doi.org/10.18653/v1/2020.acl-main.197} {{SMART}:
  Robust and efficient fine-tuning for pre-trained natural language models
  through principled regularized optimization}.
\newblock In \emph{Proceedings of the 58th Annual Meeting of the Association
  for Computational Linguistics}, pages 2177--2190, Online. Association for
  Computational Linguistics.

\bibitem[{John et~al.(2019)John, Mou, Bahuleyan, and
  Vechtomova}]{john-etal-2019-disentangled}
Vineet John, Lili Mou, Hareesh Bahuleyan, and Olga Vechtomova. 2019.
\newblock \href {https://doi.org/10.18653/v1/P19-1041} {Disentangled
  representation learning for non-parallel text style transfer}.
\newblock In \emph{Proceedings of the 57th Annual Meeting of the Association
  for Computational Linguistics}, pages 424--434, Florence, Italy. Association
  for Computational Linguistics.

\bibitem[{Krishna et~al.(2020)Krishna, Wieting, and
  Iyyer}]{krishna-etal-2020-reformulating}
Kalpesh Krishna, John Wieting, and Mohit Iyyer. 2020.
\newblock \href {https://doi.org/10.18653/v1/2020.emnlp-main.55} {Reformulating
  unsupervised style transfer as paraphrase generation}.
\newblock In \emph{Proceedings of the 2020 Conference on Empirical Methods in
  Natural Language Processing (EMNLP)}, pages 737--762, Online. Association for
  Computational Linguistics.

\bibitem[{Lewis et~al.(2020)Lewis, Liu, Goyal, Ghazvininejad, Mohamed, Levy,
  Stoyanov, and Zettlemoyer}]{lewis-etal-2020-bart}
Mike Lewis, Yinhan Liu, Naman Goyal, Marjan Ghazvininejad, Abdelrahman Mohamed,
  Omer Levy, Veselin Stoyanov, and Luke Zettlemoyer. 2020.
\newblock \href {https://doi.org/10.18653/v1/2020.acl-main.703} {{BART}:
  Denoising sequence-to-sequence pre-training for natural language generation,
  translation, and comprehension}.
\newblock In \emph{Proceedings of the 58th Annual Meeting of the Association
  for Computational Linguistics}, pages 7871--7880, Online. Association for
  Computational Linguistics.

\bibitem[{Li et~al.(2018)Li, Jia, He, and Liang}]{li-etal-2018-delete}
Juncen Li, Robin Jia, He~He, and Percy Liang. 2018.
\newblock \href {https://doi.org/10.18653/v1/N18-1169} {Delete, retrieve,
  generate: a simple approach to sentiment and style transfer}.
\newblock In \emph{Proceedings of the 2018 Conference of the North {A}merican
  Chapter of the Association for Computational Linguistics: Human Language
  Technologies, Volume 1 (Long Papers)}, pages 1865--1874, New Orleans,
  Louisiana. Association for Computational Linguistics.

\bibitem[{Lin et~al.(2021)Lin, Gao, Yan, Moreno, and
  Ren}]{lin-etal-2021-rockner}
Bill~Yuchen Lin, Wenyang Gao, Jun Yan, Ryan Moreno, and Xiang Ren. 2021.
\newblock \href {https://doi.org/10.18653/v1/2021.emnlp-main.302} {{R}ock{NER}:
  A simple method to create adversarial examples for evaluating the robustness
  of named entity recognition models}.
\newblock In \emph{Proceedings of the 2021 Conference on Empirical Methods in
  Natural Language Processing}, pages 3728--3737, Online and Punta Cana,
  Dominican Republic. Association for Computational Linguistics.

\bibitem[{Liu et~al.(2021{\natexlab{a}})Liu, Ding, Bing, Joty, Si, and
  Miao}]{liu-etal-2021-mulda}
Linlin Liu, Bosheng Ding, Lidong Bing, Shafiq Joty, Luo Si, and Chunyan Miao.
  2021{\natexlab{a}}.
\newblock \href {https://doi.org/10.18653/v1/2021.acl-long.453} {{M}ul{DA}: A
  multilingual data augmentation framework for low-resource cross-lingual
  {NER}}.
\newblock In \emph{Proceedings of the 59th Annual Meeting of the Association
  for Computational Linguistics and the 11th International Joint Conference on
  Natural Language Processing (Volume 1: Long Papers)}, pages 5834--5846,
  Online. Association for Computational Linguistics.

\bibitem[{Liu et~al.(2021{\natexlab{b}})Liu, Neubig, and
  Wieting}]{liu-etal-2021-learning}
Yixin Liu, Graham Neubig, and John Wieting. 2021{\natexlab{b}}.
\newblock \href {https://doi.org/10.18653/v1/2021.naacl-main.337} {On learning
  text style transfer with direct rewards}.
\newblock In \emph{Proceedings of the 2021 Conference of the North American
  Chapter of the Association for Computational Linguistics: Human Language
  Technologies}, pages 4262--4273, Online. Association for Computational
  Linguistics.

\bibitem[{Lu et~al.(2021)Lu, Lin, Xu, Han, Tang, Li, Sun, Liao, and
  Chen}]{lu-etal-2021-text2event}
Yaojie Lu, Hongyu Lin, Jin Xu, Xianpei Han, Jialong Tang, Annan Li, Le~Sun,
  Meng Liao, and Shaoyi Chen. 2021.
\newblock \href {https://doi.org/10.18653/v1/2021.acl-long.217}
  {{T}ext2{E}vent: Controllable sequence-to-structure generation for end-to-end
  event extraction}.
\newblock In \emph{Proceedings of the 59th Annual Meeting of the Association
  for Computational Linguistics and the 11th International Joint Conference on
  Natural Language Processing (Volume 1: Long Papers)}, pages 2795--2806,
  Online. Association for Computational Linguistics.

\bibitem[{Ma et~al.(2020)Ma, Sap, Rashkin, and
  Choi}]{ma-etal-2020-powertransformer}
Xinyao Ma, Maarten Sap, Hannah Rashkin, and Yejin Choi. 2020.
\newblock \href {https://doi.org/10.18653/v1/2020.emnlp-main.602}
  {{P}ower{T}ransformer: Unsupervised controllable revision for biased language
  correction}.
\newblock In \emph{Proceedings of the 2020 Conference on Empirical Methods in
  Natural Language Processing (EMNLP)}, pages 7426--7441, Online. Association
  for Computational Linguistics.

\bibitem[{Malmi et~al.(2020)Malmi, Severyn, and
  Rothe}]{malmi-etal-2020-unsupervised}
Eric Malmi, Aliaksei Severyn, and Sascha Rothe. 2020.
\newblock \href {https://doi.org/10.18653/v1/2020.emnlp-main.699} {Unsupervised
  text style transfer with padded masked language models}.
\newblock In \emph{Proceedings of the 2020 Conference on Empirical Methods in
  Natural Language Processing (EMNLP)}, pages 8671--8680, Online. Association
  for Computational Linguistics.

\bibitem[{Maynez et~al.(2020)Maynez, Narayan, Bohnet, and
  McDonald}]{maynez-etal-2020-faithfulness}
Joshua Maynez, Shashi Narayan, Bernd Bohnet, and Ryan McDonald. 2020.
\newblock \href {https://doi.org/10.18653/v1/2020.acl-main.173} {On
  faithfulness and factuality in abstractive summarization}.
\newblock In \emph{Proceedings of the 58th Annual Meeting of the Association
  for Computational Linguistics}, pages 1906--1919, Online. Association for
  Computational Linguistics.

\bibitem[{Morris et~al.(2020)Morris, Lifland, Yoo, Grigsby, Jin, and
  Qi}]{morris-etal-2020-textattack}
John Morris, Eli Lifland, Jin~Yong Yoo, Jake Grigsby, Di~Jin, and Yanjun Qi.
  2020.
\newblock \href {https://doi.org/10.18653/v1/2020.emnlp-demos.16}
  {{T}ext{A}ttack: A framework for adversarial attacks, data augmentation, and
  adversarial training in {NLP}}.
\newblock In \emph{Proceedings of the 2020 Conference on Empirical Methods in
  Natural Language Processing: System Demonstrations}, pages 119--126, Online.
  Association for Computational Linguistics.

\bibitem[{Nie et~al.(2020)Nie, Tian, Wan, Song, and Dai}]{nie-etal-2020-named}
Yuyang Nie, Yuanhe Tian, Xiang Wan, Yan Song, and Bo~Dai. 2020.
\newblock \href {https://doi.org/10.18653/v1/2020.emnlp-main.107} {Named entity
  recognition for social media texts with semantic augmentation}.
\newblock In \emph{Proceedings of the 2020 Conference on Empirical Methods in
  Natural Language Processing (EMNLP)}, pages 1383--1391, Online. Association
  for Computational Linguistics.

\bibitem[{Niu and Bansal(2018)}]{niu-bansal-2018-polite}
Tong Niu and Mohit Bansal. 2018.
\newblock \href {https://doi.org/10.1162/tacl_a_00027} {Polite dialogue
  generation without parallel data}.
\newblock \emph{Transactions of the Association for Computational Linguistics},
  6:373--389.

\bibitem[{Niu et~al.(2018)Niu, Rao, and Carpuat}]{niu-etal-2018-multi}
Xing Niu, Sudha Rao, and Marine Carpuat. 2018.
\newblock \href {https://aclanthology.org/C18-1086} {Multi-task neural models
  for translating between styles within and across languages}.
\newblock In \emph{Proceedings of the 27th International Conference on
  Computational Linguistics}, pages 1008--1021, Santa Fe, New Mexico, USA.
  Association for Computational Linguistics.

\bibitem[{Prabhumoye et~al.(2018)Prabhumoye, Tsvetkov, Salakhutdinov, and
  Black}]{prabhumoye-etal-2018-style}
Shrimai Prabhumoye, Yulia Tsvetkov, Ruslan Salakhutdinov, and Alan~W Black.
  2018.
\newblock \href {https://doi.org/10.18653/v1/P18-1080} {Style transfer through
  back-translation}.
\newblock In \emph{Proceedings of the 56th Annual Meeting of the Association
  for Computational Linguistics (Volume 1: Long Papers)}, pages 866--876,
  Melbourne, Australia. Association for Computational Linguistics.

\bibitem[{Provilkov et~al.(2020)Provilkov, Emelianenko, and
  Voita}]{provilkov-etal-2020-bpe}
Ivan Provilkov, Dmitrii Emelianenko, and Elena Voita. 2020.
\newblock \href {https://doi.org/10.18653/v1/2020.acl-main.170} {{BPE}-dropout:
  Simple and effective subword regularization}.
\newblock In \emph{Proceedings of the 58th Annual Meeting of the Association
  for Computational Linguistics}, pages 1882--1892, Online. Association for
  Computational Linguistics.

\bibitem[{Qi et~al.(2021)Qi, Chen, Zhang, Li, Liu, and Sun}]{qi-etal-2021-mind}
Fanchao Qi, Yangyi Chen, Xurui Zhang, Mukai Li, Zhiyuan Liu, and Maosong Sun.
  2021.
\newblock \href {https://doi.org/10.18653/v1/2021.emnlp-main.374} {Mind the
  style of text! adversarial and backdoor attacks based on text style
  transfer}.
\newblock In \emph{Proceedings of the 2021 Conference on Empirical Methods in
  Natural Language Processing}, pages 4569--4580, Online and Punta Cana,
  Dominican Republic. Association for Computational Linguistics.

\bibitem[{Raffel et~al.(2020)Raffel, Shazeer, Roberts, Lee, Narang, Matena,
  Zhou, Li, and Liu}]{2020t5}
Colin Raffel, Noam Shazeer, Adam Roberts, Katherine Lee, Sharan Narang, Michael
  Matena, Yanqi Zhou, Wei Li, and Peter~J. Liu. 2020.
\newblock \href {http://jmlr.org/papers/v21/20-074.html} {Exploring the limits
  of transfer learning with a unified text-to-text transformer}.
\newblock \emph{Journal of Machine Learning Research}, 21(140):1--67.

\bibitem[{Raman et~al.(2022)Raman, Naim, Chen, Hashimoto, Yalasangi, and
  Srinivasan}]{Raman2022TransformingST}
Karthik Raman, Iftekhar Naim, Jiecao Chen, Kazuma Hashimoto, Kiran Yalasangi,
  and Krishna Srinivasan. 2022.
\newblock \href {https://arxiv.org/abs/2203.08378} {Transforming sequence
  tagging into a seq2seq task}.
\newblock \emph{ArXiv}, abs/2203.08378.

\bibitem[{Rao and Tetreault(2018)}]{rao-tetreault-2018-dear}
Sudha Rao and Joel Tetreault. 2018.
\newblock \href {https://doi.org/10.18653/v1/N18-1012} {Dear sir or madam, may
  {I} introduce the {GYAFC} dataset: Corpus, benchmarks and metrics for
  formality style transfer}.
\newblock In \emph{Proceedings of the 2018 Conference of the North {A}merican
  Chapter of the Association for Computational Linguistics: Human Language
  Technologies, Volume 1 (Long Papers)}, pages 129--140, New Orleans,
  Louisiana. Association for Computational Linguistics.

\bibitem[{Rijhwani and
  Preotiuc-Pietro(2020)}]{rijhwani-preotiuc-pietro-2020-temporally}
Shruti Rijhwani and Daniel Preotiuc-Pietro. 2020.
\newblock \href {https://doi.org/10.18653/v1/2020.acl-main.680}
  {Temporally-informed analysis of named entity recognition}.
\newblock In \emph{Proceedings of the 58th Annual Meeting of the Association
  for Computational Linguistics}, pages 7605--7617, Online. Association for
  Computational Linguistics.

\bibitem[{Shen et~al.(2017)Shen, Lei, Barzilay, and
  Jaakkola}]{DBLP:conf/nips/ShenLBJ17}
Tianxiao Shen, Tao Lei, Regina Barzilay, and Tommi~S. Jaakkola. 2017.
\newblock \href
  {https://proceedings.neurips.cc/paper/2017/hash/2d2c8394e31101a261abf1784302bf75-Abstract.html}
  {Style transfer from non-parallel text by cross-alignment}.
\newblock In \emph{Advances in Neural Information Processing Systems 30: Annual
  Conference on Neural Information Processing Systems 2017, December 4-9, 2017,
  Long Beach, CA, {USA}}, pages 6830--6841.

\bibitem[{Simoncini and Spanakis(2021)}]{simoncini-spanakis-2021-seqattack}
Walter Simoncini and Gerasimos Spanakis. 2021.
\newblock \href {https://doi.org/10.18653/v1/2021.emnlp-demo.35}
  {{S}eq{A}ttack: {O}n adversarial attacks for named entity recognition}.
\newblock In \emph{Proceedings of the 2021 Conference on Empirical Methods in
  Natural Language Processing: System Demonstrations}, pages 308--318, Online
  and Punta Cana, Dominican Republic. Association for Computational
  Linguistics.

\bibitem[{Tjong Kim~Sang and
  Veenstra(1999)}]{tjong-kim-sang-veenstra-1999-representing}
Erik~F. Tjong Kim~Sang and Jorn Veenstra. 1999.
\newblock \href {https://aclanthology.org/E99-1023} {Representing text chunks}.
\newblock In \emph{Ninth Conference of the {E}uropean Chapter of the
  Association for Computational Linguistics}, pages 173--179, Bergen, Norway.
  Association for Computational Linguistics.

\bibitem[{Vaswani et~al.(2017)Vaswani, Shazeer, Parmar, Uszkoreit, Jones,
  Gomez, Kaiser, and Polosukhin}]{DBLP:conf/nips/VaswaniSPUJGKP17}
Ashish Vaswani, Noam Shazeer, Niki Parmar, Jakob Uszkoreit, Llion Jones,
  Aidan~N. Gomez, Lukasz Kaiser, and Illia Polosukhin. 2017.
\newblock \href
  {https://proceedings.neurips.cc/paper/2017/hash/3f5ee243547dee91fbd053c1c4a845aa-Abstract.html}
  {Attention is all you need}.
\newblock In \emph{Advances in Neural Information Processing Systems 30: Annual
  Conference on Neural Information Processing Systems 2017, December 4-9, 2017,
  Long Beach, CA, {USA}}, pages 5998--6008.

\bibitem[{Wang and Henao(2021)}]{wang-henao-2021-unsupervised}
Rui Wang and Ricardo Henao. 2021.
\newblock \href {https://doi.org/10.18653/v1/2021.emnlp-main.430} {Unsupervised
  paraphrasing consistency training for low resource named entity recognition}.
\newblock In \emph{Proceedings of the 2021 Conference on Empirical Methods in
  Natural Language Processing}, pages 5303--5308, Online and Punta Cana,
  Dominican Republic. Association for Computational Linguistics.

\bibitem[{Wang et~al.(2019)Wang, Wu, Mou, Li, and
  Chao}]{wang-etal-2019-harnessing}
Yunli Wang, Yu~Wu, Lili Mou, Zhoujun Li, and Wenhan Chao. 2019.
\newblock \href {https://doi.org/10.18653/v1/D19-1365} {Harnessing pre-trained
  neural networks with rules for formality style transfer}.
\newblock In \emph{Proceedings of the 2019 Conference on Empirical Methods in
  Natural Language Processing and the 9th International Joint Conference on
  Natural Language Processing (EMNLP-IJCNLP)}, pages 3573--3578, Hong Kong,
  China. Association for Computational Linguistics.

\bibitem[{Wang et~al.(2020)Wang, Wu, Mou, Li, and
  Chao}]{wang-etal-2020-formality}
Yunli Wang, Yu~Wu, Lili Mou, Zhoujun Li, and WenHan Chao. 2020.
\newblock \href {https://doi.org/10.18653/v1/2020.coling-main.203} {Formality
  style transfer with shared latent space}.
\newblock In \emph{Proceedings of the 28th International Conference on
  Computational Linguistics}, pages 2236--2249, Barcelona, Spain (Online).
  International Committee on Computational Linguistics.

\bibitem[{Wang et~al.(2018)Wang, Qu, Chen, Shen, Zhang, Zhang, Gao, Gu, Chen,
  and Yu}]{wang-etal-2018-label-aware}
Zhenghui Wang, Yanru Qu, Liheng Chen, Jian Shen, Weinan Zhang, Shaodian Zhang,
  Yimei Gao, Gen Gu, Ken Chen, and Yong Yu. 2018.
\newblock \href {https://doi.org/10.18653/v1/N18-1001} {Label-aware double
  transfer learning for cross-specialty medical named entity recognition}.
\newblock In \emph{Proceedings of the 2018 Conference of the North {A}merican
  Chapter of the Association for Computational Linguistics: Human Language
  Technologies, Volume 1 (Long Papers)}, pages 1--15, New Orleans, Louisiana.
  Association for Computational Linguistics.

\bibitem[{Wei and Zou(2019)}]{wei-zou-2019-eda}
Jason Wei and Kai Zou. 2019.
\newblock \href {https://doi.org/10.18653/v1/D19-1670} {{EDA}: Easy data
  augmentation techniques for boosting performance on text classification
  tasks}.
\newblock In \emph{Proceedings of the 2019 Conference on Empirical Methods in
  Natural Language Processing and the 9th International Joint Conference on
  Natural Language Processing (EMNLP-IJCNLP)}, pages 6382--6388, Hong Kong,
  China. Association for Computational Linguistics.

\bibitem[{Welleck et~al.(2020)Welleck, Kulikov, Roller, Dinan, Cho, and
  Weston}]{DBLP:conf/iclr/WelleckKRDCW20}
Sean Welleck, Ilia Kulikov, Stephen Roller, Emily Dinan, Kyunghyun Cho, and
  Jason Weston. 2020.
\newblock \href {https://openreview.net/forum?id=SJeYe0NtvH} {Neural text
  generation with unlikelihood training}.
\newblock In \emph{8th International Conference on Learning Representations,
  {ICLR} 2020, Addis Ababa, Ethiopia, April 26-30, 2020}. OpenReview.net.

\bibitem[{Wenjing et~al.(2021)Wenjing, Jian, Jinan, Yufeng, and
  Yujie}]{wenjing-etal-2021-improving}
Zhu Wenjing, Liu Jian, Xu~Jinan, Chen Yufeng, and Zhang Yujie. 2021.
\newblock \href {https://aclanthology.org/2021.ccl-1.101} {Improving
  low-resource named entity recognition via label-aware data augmentation and
  curriculum denoising}.
\newblock In \emph{Proceedings of the 20th Chinese National Conference on
  Computational Linguistics}, pages 1131--1142, Huhhot, China. Chinese
  Information Processing Society of China.

\bibitem[{Wolf et~al.(2020)Wolf, Debut, Sanh, Chaumond, Delangue, Moi, Cistac,
  Rault, Louf, Funtowicz, Davison, Shleifer, von Platen, Ma, Jernite, Plu, Xu,
  Le~Scao, Gugger, Drame, Lhoest, and Rush}]{wolf-etal-2020-transformers}
Thomas Wolf, Lysandre Debut, Victor Sanh, Julien Chaumond, Clement Delangue,
  Anthony Moi, Pierric Cistac, Tim Rault, Remi Louf, Morgan Funtowicz, Joe
  Davison, Sam Shleifer, Patrick von Platen, Clara Ma, Yacine Jernite, Julien
  Plu, Canwen Xu, Teven Le~Scao, Sylvain Gugger, Mariama Drame, Quentin Lhoest,
  and Alexander Rush. 2020.
\newblock \href {https://doi.org/10.18653/v1/2020.emnlp-demos.6} {Transformers:
  State-of-the-art natural language processing}.
\newblock In \emph{Proceedings of the 2020 Conference on Empirical Methods in
  Natural Language Processing: System Demonstrations}, pages 38--45, Online.
  Association for Computational Linguistics.

\bibitem[{Xia et~al.(2019)Xia, Kong, Anastasopoulos, and
  Neubig}]{xia-etal-2019-generalized}
Mengzhou Xia, Xiang Kong, Antonios Anastasopoulos, and Graham Neubig. 2019.
\newblock \href {https://doi.org/10.18653/v1/P19-1579} {Generalized data
  augmentation for low-resource translation}.
\newblock In \emph{Proceedings of the 57th Annual Meeting of the Association
  for Computational Linguistics}, pages 5786--5796, Florence, Italy.
  Association for Computational Linguistics.

\bibitem[{Yamada et~al.(2020)Yamada, Asai, Shindo, Takeda, and
  Matsumoto}]{yamada-etal-2020-luke}
Ikuya Yamada, Akari Asai, Hiroyuki Shindo, Hideaki Takeda, and Yuji Matsumoto.
  2020.
\newblock \href {https://doi.org/10.18653/v1/2020.emnlp-main.523} {{LUKE}: Deep
  contextualized entity representations with entity-aware self-attention}.
\newblock In \emph{Proceedings of the 2020 Conference on Empirical Methods in
  Natural Language Processing (EMNLP)}, pages 6442--6454, Online. Association
  for Computational Linguistics.

\bibitem[{Yang et~al.(2019)Yang, Dai, Yang, Carbonell, Salakhutdinov, and
  Le}]{DBLP:conf/nips/YangDYCSL19}
Zhilin Yang, Zihang Dai, Yiming Yang, Jaime~G. Carbonell, Ruslan Salakhutdinov,
  and Quoc~V. Le. 2019.
\newblock \href
  {https://proceedings.neurips.cc/paper/2019/hash/dc6a7e655d7e5840e66733e9ee67cc69-Abstract.html}
  {Xlnet: Generalized autoregressive pretraining for language understanding}.
\newblock In \emph{Advances in Neural Information Processing Systems 32: Annual
  Conference on Neural Information Processing Systems 2019, NeurIPS 2019,
  December 8-14, 2019, Vancouver, BC, Canada}, pages 5754--5764.

\bibitem[{Yang et~al.(2018)Yang, Hu, Dyer, Xing, and
  Berg{-}Kirkpatrick}]{DBLP:conf/nips/YangHDXB18}
Zichao Yang, Zhiting Hu, Chris Dyer, Eric~P. Xing, and Taylor
  Berg{-}Kirkpatrick. 2018.
\newblock \href
  {https://proceedings.neurips.cc/paper/2018/hash/398475c83b47075e8897a083e97eb9f0-Abstract.html}
  {Unsupervised text style transfer using language models as discriminators}.
\newblock In \emph{Advances in Neural Information Processing Systems 31: Annual
  Conference on Neural Information Processing Systems 2018, NeurIPS 2018,
  December 3-8, 2018, Montr{\'{e}}al, Canada}, pages 7298--7309.

\bibitem[{Zaheer et~al.(2020)Zaheer, Guruganesh, Dubey, Ainslie, Alberti,
  Onta{\~{n}}{\'{o}}n, Pham, Ravula, Wang, Yang, and
  Ahmed}]{DBLP:conf/nips/ZaheerGDAAOPRWY20}
Manzil Zaheer, Guru Guruganesh, Kumar~Avinava Dubey, Joshua Ainslie, Chris
  Alberti, Santiago Onta{\~{n}}{\'{o}}n, Philip Pham, Anirudh Ravula, Qifan
  Wang, Li~Yang, and Amr Ahmed. 2020.
\newblock \href
  {https://proceedings.neurips.cc/paper/2020/hash/c8512d142a2d849725f31a9a7a361ab9-Abstract.html}
  {Big bird: Transformers for longer sequences}.
\newblock In \emph{Advances in Neural Information Processing Systems 33: Annual
  Conference on Neural Information Processing Systems 2020, NeurIPS 2020,
  December 6-12, 2020, virtual}.

\bibitem[{Zeng et~al.(2020)Zeng, Li, Zhai, and
  Zhang}]{zeng-etal-2020-counterfactual}
Xiangji Zeng, Yunliang Li, Yuchen Zhai, and Yin Zhang. 2020.
\newblock \href {https://doi.org/10.18653/v1/2020.emnlp-main.590}
  {Counterfactual generator: A weakly-supervised method for named entity
  recognition}.
\newblock In \emph{Proceedings of the 2020 Conference on Empirical Methods in
  Natural Language Processing (EMNLP)}, pages 7270--7280, Online. Association
  for Computational Linguistics.

\bibitem[{Zhang et~al.(2021)Zhang, Xia, Yu, Liu, and
  Zhao}]{zhang-etal-2021-pdaln}
Tao Zhang, Congying Xia, Philip~S. Yu, Zhiwei Liu, and Shu Zhao. 2021.
\newblock \href {https://doi.org/10.18653/v1/2021.emnlp-main.442} {{PDALN}:
  Progressive domain adaptation over a pre-trained model for low-resource
  cross-domain named entity recognition}.
\newblock In \emph{Proceedings of the 2021 Conference on Empirical Methods in
  Natural Language Processing}, pages 5441--5451, Online and Punta Cana,
  Dominican Republic. Association for Computational Linguistics.

\bibitem[{Zhao et~al.(2021)Zhao, Ding, and Feng}]{zhao-etal-2021-glara}
Xinyan Zhao, Haibo Ding, and Zhe Feng. 2021.
\newblock \href {https://doi.org/10.18653/v1/2021.eacl-main.318} {{GL}a{RA}:
  Graph-based labeling rule augmentation for weakly supervised named entity
  recognition}.
\newblock In \emph{Proceedings of the 16th Conference of the European Chapter
  of the Association for Computational Linguistics: Main Volume}, pages
  3636--3649, Online. Association for Computational Linguistics.

\bibitem[{Zhu et~al.(2021)Zhu, Zhang, Liu, and Wang}]{zhu-etal-2021-neural}
Qingfu Zhu, Wei-Nan Zhang, Ting Liu, and William~Yang Wang. 2021.
\newblock \href {https://doi.org/10.18653/v1/2021.acl-long.339} {Neural
  stylistic response generation with disentangled latent variables}.
\newblock In \emph{Proceedings of the 59th Annual Meeting of the Association
  for Computational Linguistics and the 11th International Joint Conference on
  Natural Language Processing (Volume 1: Long Papers)}, pages 4391--4401,
  Online. Association for Computational Linguistics.

\end{thebibliography}
\bibliographystyle{acl_natbib}

\appendix

\section{Data Statistics}
\label{sec: data_statistics}
Table \ref{tab: data_statistics} presents the data statistics of GYAFC corpus \citep{rao-tetreault-2018-dear}, OntoNotes 5.0 corpus and Temporal Twitter corpus \citep{rijhwani-preotiuc-pietro-2020-temporally}. We consider totally 18 different entity types following the annotation schema of OntoNotes 5.0 corpus, including \textit{WORK\_OF\_ART}, \textit{ORG}, \textit{FAC}, \textit{LAW}, \textit{PERCENT}, \textit{PRODUCT}, \textit{MONEY}, \textit{DATE}, \textit{PERSON}, \textit{GPE}, \textit{QUANTITY}, \textit{CARDINAL}, \textit{NORP}, \textit{TIME}, \textit{EVENT}, \textit{ORDINAL}, \textit{LOC}, \textit{LANGUAGE}.

\begin{table}[ht]
    \centering
    \renewcommand{\arraystretch}{1.2}
    \resizebox{\linewidth}{!}{
    \begin{tabular}{l|c|rrr}
    \hline
    \bf Dataset & \bf Domain & \bf Train & \bf Dev & \bf Test \\
    \hline
    \bf GYAFC Corpus                            & Formality & 10,4562 & 10,268 & 4,849 \\
    \hline
    \multirow{5}{*}{\bf OntoNotes 5.0 Corpus}   & BC        & 11,879 & 2,117 & 2,211 \\
                                                & BN        & 10,683 & 1,295 & 1,357 \\
                                                & MZ        &  6,911 &   642 &   780 \\
                                                & NW        & 34,970 & 5,896 & 2,327 \\
                                                & WB        & 15,554 & 2,316 & 2,307 \\
    \hline
    \bf Temporal Twitter Corpus                 & SM & 10,000 &   500 & 1,500 \\
    \hline
    \end{tabular}
    }
    \caption{Data Statistics of GYAFC corpus, OntoNotes 5.0 corpus and Temporal Twitter corpus. }
    \label{tab: data_statistics}
\end{table}

\section{Data Preprocessing and Filtering}
\label{sec: data_preprocessing_and_filtering}
Due to the limitation of computational resources, we set a max length of 64 to filter out long linearized sentences in both style transfer dataset $\mathcal{P}$ and NER dataset $\mathcal{D}$ for training the proposed framework to generate pseudo data. We also uses a pre-trained $\text{BERT}_{\text{large}}$ model \citep{devlin-etal-2019-bert} to assign pseudo NER tags for sentences in the style transfer dataset $\mathcal{P}$ as weak supervision. The pre-trained $\text{BERT}_{\text{large}}$ model is only trained with the source data and has no access to the target data. The predicted NER tags are selected only if it comes with a confidence score (i.e., predicted probability) higher than 0.9 in both parallel source and target sentence. This results in approximately 10\% of sentences in the style transfer dataset $\mathcal{P}$ having pseudo NER tags. For the NER dataset $\mathcal{D}$, we simply adapt original tags without further providing pseudo labels.

\section{Score Calculation in Data Selection}
\label{sec: score_calculation_in_data_selection}
We simply fine-tune a $\text{T5}_{\text{base}}$ model \citep{2020t5} for style classification as the style classifier to obtain the consistency score. The adequacy and fluency scores are obtained from the softmax confidence a pretrained NLU model \footnote{\url{https://github.com/PrithivirajDamodaran/Parrot_Paraphraser}}.

\section{Hyper-parameters and Fine-tuning}
\label{sec: hyperparameters_and_finetuning}
Table \ref{tab: hyper_parameters} lists the hyper-parameters for both style transfer and NER tasks. All hyper-parameters are kept the same across different experiments for fine-tuning and/or generating. For the hardware, we use 8 NVIDIA V100 GPUs with a memory of 24GB. By adjusting the training batch size, our experiments should be compatible with any GPU that has a memory higher than 10GB.

\begin{table}[ht]
    \centering
    \renewcommand{\arraystretch}{1.2}
    \resizebox{\linewidth}{!}{
    \begin{tabular}{lccc}
    \hline
    \multirow{2}{*}{\bf Parameter}  & \multicolumn{2}{c}{\bf Style Transfer}    & \bf NER \\
    \cmidrule(lr){2-3} \cmidrule(lr){4-4}
                                    & \bf Fine-tuning   & \bf Generating        & \bf Fine-tuning \\
    \hline
    Model Type                      & $\text{T5}_{\text{base}}$ & $\text{T5}_{\text{base}}$ & $\text{BERT}_{\text{base}}$ \\
    Optimizer                       & AdamW                     & -                         & Adam \\
    Learning Rate                   & $1e-4$                    & -                         & $5e-5$ \\
    Training Epochs                 & 5                         & -                         & 10 \\
    Batch Size                      & 6                         & -                         & 32 \\
    Weight Decay                    & 0.01                      & -                         & 0.01 \\
    Top-k                           & 50                        & 50                        & - \\
    Top-p                           & 0.98                      & 0.98                      & - \\
    $\lambda_{\text{pg}}$           & 1.0                       & -                         & - \\
    $\lambda_{\text{cr}}$           & 0.5                       & -                         & - \\
    $\lambda_{\text{adv}}$          & 1.25                      & -                         & - \\
    $\tau$                          & 1.0                       & 1.5                       & - \\
    $\lambda_{\text{Consistency}}$  & -                         & 1.0                       & - \\
    $\lambda_{\text{Adequacy}}$     & -                         & 1.0                       & - \\
    $\lambda_{\text{Fluency}}$      & -                         & 0.1                       & - \\
    $\lambda_{\text{Diversity}}$    & -                         & 0.5                       & - \\
    \hline
    \end{tabular}
    }
    \caption{Hyper-parameters for style transfer and NER tasks.}
    \label{tab: hyper_parameters}
\end{table}
\begin{table*}[t!]
\resizebox{\linewidth}{!}{
\begin{tabular}[t]{@{}llll@{}}
\toprule
\textbf{\#} 
    & \textbf{Original Sentences (formal)} 
    & \textbf{Generated Sentences (informal)} 
\\\toprule

\multicolumn{3}{l}{\bf \textit{Good Examples}}
\\\toprule

1 & 
\begin{tabular}[t]{@{}p{0.55\linewidth}@{}}
    The \textbf{{[}IRS{]}\sub{ORGANIZATION}} denied cost deductions because few of the temps got to go aboard.
\end{tabular} & 
\begin{tabular}[t]{@{}p{0.55\linewidth}@{}}
    the \textbf{{[}IRS{]}\sub{ORGANIZATION}} denying cost depossions... bcoz few of the guys got to go on board
\end{tabular}
\\\midrule

2 & 
\begin{tabular}[t]{@{}p{0.55\linewidth}@{}}
    Oh yes, the \textbf{{[}Atlantic{]}\sub{ORGANIZATION}} was also pushed back \textbf{{[}300 yards{]}\sub{QUANTITY}}.
\end{tabular} & 
\begin{tabular}[t]{@{}p{0.55\linewidth}@{}}
    Yea-huh the \textbf{{[}Atlantic{]}\sub{ORGANIZATION}} was pushed back of \textbf{{[}300 feet{]}\sub{QUANTITY}} \includegraphics[scale=0.07]{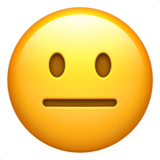}
\end{tabular}
\\\midrule

3 & 
\begin{tabular}[t]{@{}p{0.55\linewidth}@{}}
    Her neighbour, \textbf{{[}Lin{]}\sub{PERSON}}, laughs in agreement, ``Tea processing, it just kills you!"
\end{tabular} & 
\begin{tabular}[t]{@{}p{0.55\linewidth}@{}}
    Her friend \textbf{{[}Alexander Lu{]}\sub{PERSON}} likes it the same way but the time that Tea process kills you is very bad lol
\end{tabular}
\\\midrule

4 & 
\begin{tabular}[t]{@{}p{0.55\linewidth}@{}}
    Thus although he is without employment, he can while away his time by collecting stones and studying \textbf{{[}the Book of Changes{]}\sub{WORK\_OF\_ART}} and fengshui.
\end{tabular} &
\begin{tabular}[t]{@{}p{0.55\linewidth}@{}}
    So as if he has no jobs now he can keep his time by collection of jags and studying \textbf{{[}the Yingko Ceramics Museum's{]}\sub{ORGANIZATION}} and fengshui. 
\end{tabular} 
\\\midrule

5 & 
\begin{tabular}[t]{@{}p{0.55\linewidth}@{}}
    \textbf{{[}Ting{]}\sub{PERSON}} concludes : ``He has lost his direction in terms of education policy, and he underestimates the education community."
\end{tabular} & 
\begin{tabular}[t]{@{}p{0.55\linewidth}@{}}
    \textbf{{[}Ting{]}\sub{PERSON}} : He's lost his plan for education
\end{tabular} 

\\\toprule
\multicolumn{3}{l}{\bf \textit{Bad Examples}}
\\\toprule

6 & 
\begin{tabular}[t]{@{}p{0.55\linewidth}@{}}
    The departing Mr. \textbf{{[}Cathcart{]}\sub{PERSON}} says he has no worries about \textbf{{[}Kidder{]}\sub{ORGANIZATION}} 's future.
\end{tabular} & 
\begin{tabular}[t]{@{}p{0.55\linewidth}@{}}
    He says he has no worries
\end{tabular}
\\\midrule

7 & 
\begin{tabular}[t]{@{}p{0.55\linewidth}@{}}
    I don't think the checks are worth \$ \textbf{{[}15{]}\sub{MONEY}} apiece, '' he says.
\end{tabular} &
\begin{tabular}[t]{@{}p{0.55\linewidth}@{}}
    he says It's weird they're for \$ 15 per check 
\end{tabular} 
\\\midrule

8 & 
\begin{tabular}[t]{@{}p{0.55\linewidth}@{}}
    Well the rescue vehicle if it's required will have only \textbf{{[}four{]}\sub{CARDINAL}} people on it.
\end{tabular} & 
\begin{tabular}[t]{@{}p{0.55\linewidth}@{}}
    well if necessary the rescue car.. WILL have only four + people on it
\end{tabular}
\\\midrule

9 & 
\begin{tabular}[t]{@{}p{0.55\linewidth}@{}}
    it's going to be in small gathering places in middle \textbf{{[}America{]}\sub{GPE}} with people saying some pretty horrible things
\end{tabular} & 
\begin{tabular}[t]{@{}p{0.55\linewidth}@{}}
    Small gathering places in the middle \textbf{{[}America{]}\sub{LOC}} with people saying some pretty awful things
\end{tabular}
\\\midrule

10 & 
\begin{tabular}[t]{@{}p{0.55\linewidth}@{}}
    So \textbf{{[}al Jazeera TV{]}\sub{ORGANIZATION}} station has also adopted this structure.
\end{tabular} & 
\begin{tabular}[t]{@{}p{0.55\linewidth}@{}}
    Because \textbf{{[}al Jazeera{]}\sub{PERSON}} TV \textbf{{[}Imad{]}\sub{PERSON}} has too follow that structure
\end{tabular} 

\\\bottomrule
\end{tabular}
}
\caption{
Examples of data augmentation with data transformation for case study. The sentences are tagged with their corresponding entities in brackets (e.g., {[}IRS{]}\sub{ORGANIZATION}).
}
\label{tab: case_study}
\end{table*}

\section{Examples of Generated Data}
\label{sec: error_analysis}
Table \ref{tab: case_study} shows some examples of formal source sentences and their corresponding informal target sentences generated from our proposed method for case study.

\end{document}